\documentclass[conference]{IEEEtran}
\IEEEoverridecommandlockouts
\usepackage{cite}
\usepackage{amsmath,amssymb,amsfonts}
\usepackage{algorithmic}
\usepackage{graphicx}
\usepackage{textcomp}
\usepackage{xcolor}
\usepackage{adjustbox}
\usepackage{graphicx}
\usepackage{graphics}
\def\BibTeX{{\rm B\kern-.05em{\sc i\kern-.025em b}\kern-.08em
    T\kern-.1667em\lower.7ex\hbox{E}\kern-.125emX}}

\usepackage{caption}
\usepackage{subcaption}
\usepackage{mwe}

\usepackage{multicol}
\usepackage{multirow}
\usepackage{booktabs}
\usepackage{todonotes}
\usepackage[a-1b]{pdfx}
\usepackage{rotating}
\usepackage{tablefootnote}
\usepackage{booktabs}
\usepackage{wrapfig}

\begin{document}

\title{Using Infant Limb Movement Data to Control Small Aerial Robots\\

\author{Georgia Kouvoutsakis,$^{1}$ Elena Kokkoni,$^{2}$ and Konstantinos Karydis$^{1}$
\thanks{$^{1}$Dept. of Electrical and Computer Engineering; $^{2}$Dept. of Bioengineering, University of California, Riverside, 900 University Ave, Riverside, CA 92521, USA. Email:{\tt\footnotesize\{gkouv001, elenak, karydis\}@ucr.edu}. We gratefully acknowledge the support of the National Pediatric Rehabilitation Resource Center (C-PROGRESS). Any opinions, findings, and conclusions or recommendations expressed in this material are those of the authors and do not necessarily reflect the views of C-PROGRESS.}}
}


\maketitle

\begin{abstract}
Promoting exploratory movements through contingent feedback can positively influence motor development in infancy.
Our ongoing work gears toward the development of a robot-assisted contingency learning environment through the use of small aerial robots.
This paper examines whether aerial robots and their associated motion controllers can be used to achieve efficient and highly-responsive robot flight for our purpose.
Infant kicking kinematic data were extracted from videos and used in simulation and physical experiments with an aerial robot. 
The efficacy of two standard of practice controllers was assessed: a linear PID and a nonlinear geometric controller. 
The ability of the robot to match infant kicking trajectories was evaluated qualitatively and quantitatively via the mean squared error (to assess overall deviation from the input infant leg trajectory signals), and dynamic time warping algorithm (to quantify the signal synchrony). Results demonstrate that it is in principle possible to track infant kicking trajectories with small aerials robots, and identify areas of further development required to improve the tracking quality.
\end{abstract}


\section{Introduction}
The ability of humans to perform movements early in life sets the basis for developing more complex motor patterns later; an example is alternating kicking\cite{piek1997spontaneous}, which is suggested to be a precursor of walking \cite{jeng2004relationship, ulrich1995spontaneous}.
On the other hand, motor delays early on may have a significant impact on lifelong mobility \cite{adolph2019ecological,campos2000travel}. 
Applying early interventions can change the course of motor development \cite{novak2017early,herskind2015early}, especially if these involve stimulation of active motor exploration \cite{blauw2005systematic, angulo2001exploration}. 
The mobile contingency paradigm is an example of the latter \cite{rovee1978topographical}. 
Studies that utilize this paradigm show that infants are able to learn new motor actions by adjusting their limb movements to gain a reward from a mobile toy in the environment; the motion correspondence between the mobile toy and the infant result to alterations in rate, amplitude and coordination of infants’ kicking \cite{heathcock2004performance,thelen1983spontaneous,sargent2014infant,sargent2015development, sargent2022motivating}.
Recently, a robotic overhead mobile system was introduced to promote kicking actions \cite{emeli2020robotic}. 

Robots have been increasingly gaining presence in early intervention paradigms.
The goal is to encourage motor exploration in infants through imitation and reward-based interactions \cite{kolobe2015effectiveness,kokkoni2020gearing,kouvoutsakis2022feasibility}.  
A relevant paradigm to this paper involves the use of a humanoid robot to encourage single-leg kicking actions in infants while sitting \cite{fitter2019socially}. 
Often in these paradigms, robots need to recognize and match infant motion patterns.

Previous studies with adults have addressed this topic.
Controllers have been developed to maintain balance of humanoid robots while tracking a given reference of human motion. \cite{yamane2010controlling, ott2008motion}.
Similarly, human leg motion data have been mapped to humanoids \cite{gong2022motion,bindal2015design, gobee2017humanoid}, and leg task models have been used to achieve learning from observation in humanoid robots for dancing and stepping \cite{nakaoka2007learning}. 
In our work, we are gearing toward the deployment of small aerial robots to provide contingent rewards and promote limb movements in infants.
Currently, limited information exists on the use of aerial robots and associated motion controllers to achieve efficient and highly-responsive robot flight that can match infant limb movements.

Aerial robots allow motion in different directions, amplitudes, and velocities, which can be very useful for our paradigm that requires more complex actions from the mobile.
In general, aerial robots can interact with humans and react to gestures in a personalized manner \cite{cauchard2015drone, rajappa2017design}.
In our case, we need the aerial robot to be able to match its flight characteristics to infant motion produced during spontaneous kicking.
Previously, acquisition and processing of body signals (from adults) using body - machine interfaces have been used for the generation of control inputs for aerial robots \cite{macchini2021impact}.
Similarly, our main objective is to assess the feasibility of using an aerial robot for the mobile paradigm by determining whether the robot is able to generate infant kicking motion through the use of specific controllers. 
This is the first step leading to the development of our robot-assisted learning environment.
 
\section{Methods}

\subsection{Acquisition of Infant Kicking Trajectories}
A dataset of infant leg movements was compiled from open source videos from YouTube and publicly available videos from previous studies \cite{sargent2014infant,sargent2022motivating} using the search terms ``infant kicking" and ``infant kicking supine."
Videos from 10 infants less than six months\footnote{Age was obtained either from video descriptions (S1-4, S6, S8-9) or estimated by the authors based on displayed level of ability (S5, S7, S10).} of age were considered.
A total of 13 video segments (shown in bold font in Table~\ref{tab:subjects}) were selected based on the following inclusion criteria:
(i) infant was placed in the supine position;
(ii) infant's full body was visible and their trunk/pelvis was motionless; 
(iii) camera was stationary and placed on the side of the infant;
(iv) some known real - world object could be identified in the video to use as a calibration measure (in cases where no such object could be identified, and the infant gazed at the camera, their cornea was used for calibration~\cite{ronneburger2006growth});
(v) segment length was greater than eight seconds in an effort to obtain most kicking cycles possible (ranging from less than one to more than eight seconds\cite{thelen1981spontaneous}). 
Video segments were used to digitize a point on the ankle joint and obtain its two-dimensional position at every frame.

\begin{table}[!h]
\vspace{-6pt}
\centering
\caption{Subjects and Video segments}
\label{tab:subjects}
\resizebox{\columnwidth}{!}{%
\begin{tabular}{
c c c c}
 \toprule
 \multirow{1}{*}{Subject} &
 \multirow{1}{*}{Age} &
 \multirow{1}{*}{Segments} & Duration \\
 ID & [months] & [number] & [sec] \\
 \midrule
 S1& 3 & 1 &  \textbf{25.33 }\\
 S2& 2 & 5 & \textbf{9.98}, 6.87, 2.47, \textbf{10.24}, \textbf{9.74}
\\
 S3& 6 & 3  & 2.77, 4.67, \textbf{9.31}\\
 S4& 3.5 - 4.5 &1 & \textbf{11.87}\\
 S5& $<$ 6 &2& 6.26, \textbf{8.60}\\
 S6& 2 &1&\textbf{30.73}\\
 S7& $<$ 6  &2& 5.17, \textbf{8.17}
\\
 S8&  3 &1& \textbf{163.33}\\
 S9& 3 &2& \textbf{46.98, 9.54}
\\
 S10& $<$ 6 &1& \textbf{20.92}\\
 \bottomrule
\end{tabular}%
}
\vspace{-6pt}
\end{table}



\vspace{-12pt}
\subsection{Aerial Robot Dynamic Modeling and Control}
\begin{figure}[!h]
\vspace{-6pt}
\centering
\includegraphics[trim={0cm 5cm 18.5cm 0cm},clip,width=0.65\columnwidth]{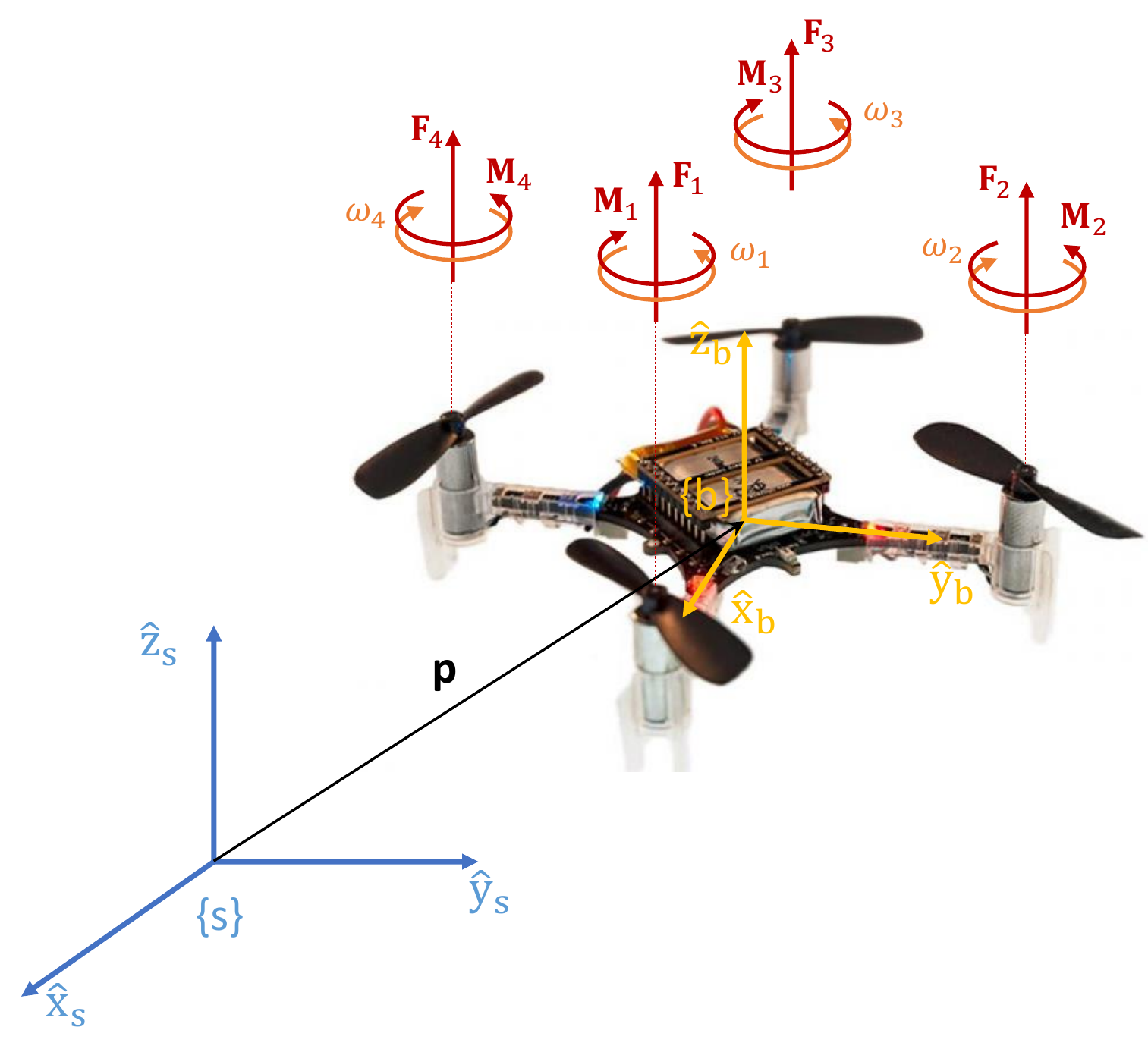}
\vspace{-6pt}
\caption{The aerial robot used herein, along with frames and annotated forces and moments. (Figure best viewed in color.)}
\label{fig:quad}
\vspace{-6pt}
\end{figure}

The Crazyflie 2.0 quadrotor aerial robot (Fig.~\ref{fig:quad}) was used. 
The selection of the specific robot was based on its small form factor and low weight (the robot weighs $27$\;g including an 1-cell $240$\;mAh battery and measures $92$\;mm W. x $92$\;mm H. x $29$\;mm D. motor-to-motor and including motor mount feet), which render it safe for use in an environment with infants.

The robot was tasked to operate in the $x-z$ plane, but robot dynamics and control were not explicitly constrained to be planar so as to facilitate future work that will focus on complete 3D flight. 
With reference to Fig.~\ref{fig:quad}, the force and moment produced by each rotor $i={1,\ldots,4}$ are given by $||{\bf F}_i||=F_i=k_f\omega_i^2$ and $||{\bf M}_i||=M_i=k_m\omega_i^2$, respectively (bold font symbols denote vectors). 
Thrust ($k_f$) and moment ($k_m$) coefficients are typically estimated experimentally; in our case we used those reported by the robot manufacturer (we used stock components only). 
Force and moment equilibrium enable computation of the total thrust and moment vectors as 
\begin{align*}
    {\bf F} &={\bf F}_1 + {\bf F}_2 + {\bf F}_3 + {\bf F}_4 - mg{\bf \hat{z}}_s\enspace,\\
    {\bf M} &={\bf M}_1 + {\bf M}_2 + {\bf M}_3 + {\bf M}_4\\
    & ~~~ + l{\bf \hat{x}}_b\times{\bf F}_1 + l{\bf \hat{y}}_b\times{\bf F}_2 - l{\bf \hat{x}}_b\times{\bf F}_3 - l{\bf \hat{y}}_b\times{\bf F}_4\enspace,
\end{align*}
where $m$ and $l$ are the mass and arm length of the robot, and $g$ denotes the gravity constant. Indices $s$ and $b$ indicate quantities expressed in the inertial $\{s\}$ and body $\{b\}$ frames, respectively; hatted symbols correspond to unit vectors.

We can then derive the dynamic equations of motion for the aerial robot as follows.\footnote{~Derivations are based on the Newton-Euler dynamics formulation. Additional details can be found in works such as~\cite{beard2008quadrotor}.} Translational dynamics is given by 
\begin{equation}\label{eq:translationalDyn}
    m{\bf \ddot{p}}_b=\begin{bmatrix}
    0\\0\\mg\end{bmatrix}+R_{sb}\begin{bmatrix}0\\0\\F_1+F_2+F_3+F_4\end{bmatrix}\enspace,
\end{equation}
with $R_{sb}$ the rotation matrix expressing the orientation of frame $\{b\}$ in terms of frame $\{s\}$. The rotational dynamics is given by 
\begin{equation}\label{eq:rotationalDyn}
    I_b\begin{bmatrix}\dot{\omega}_x\\\dot{\omega}_y\\\dot{\omega}_z\end{bmatrix}\hspace{-1pt}=\hspace{-1pt}\begin{bmatrix}l(F_2-F_4)\\l(F_3-F_1)\\M_1-M_2+M_3-M_4\end{bmatrix} - \begin{bmatrix}\omega_x\\\omega_y\\\omega_z\end{bmatrix} \times I_b\begin{bmatrix}\omega_x\\\omega_y\\\omega_z\end{bmatrix}\hspace{-3pt},
\end{equation}
where $I_b$ is the $3\times 3$ inertia matrix and ${\bf \omega}_b=[\omega_x,\omega_y,\omega_z]^T$ is the robot angular velocity expressed in the body frame.

The control input to the robot contains the thrust and moments about the three axes. Letting $\gamma=\frac{k_m}{k_f}=\frac{M_i}{F_i}$ yields $M_i=\gamma F_i$; the control input can be compactly written as
\begin{equation}\label{eq:input}
    {\bf u}=\begin{bmatrix}1 & 1 & 1 & 1\\0 & l & 0 & -l\\-l & 0 & l & 0\\\gamma & -\gamma & \gamma & -\gamma\end{bmatrix}\begin{bmatrix}F_1\\F_2\\F_3\\F_4
    \end{bmatrix}\enspace.
\end{equation}
%


\begin{figure*}[!h]
\vspace{0pt}
\centering
    \begin{subfigure}[h]{0.37\columnwidth}
        \centering
        \includegraphics[trim={10pt 5pt 40pt 35pt},clip,width=0.99\columnwidth]{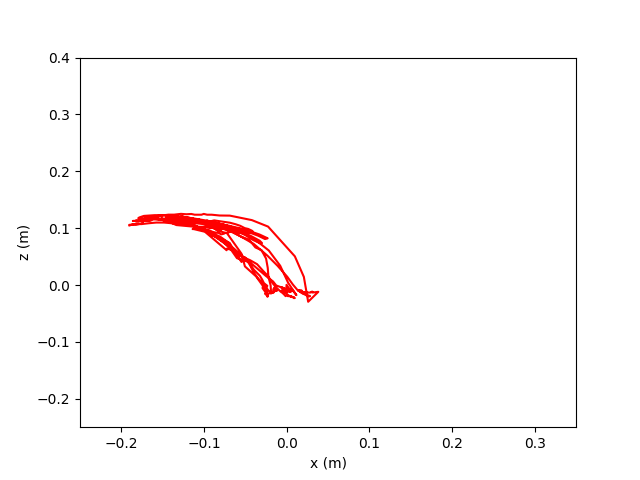}
    \end{subfigure}
    \hspace{1pt}
    \begin{subfigure}[h]{0.37\columnwidth}
        \centering
        \includegraphics[trim={10pt 5pt 40pt 35pt},clip,width=0.99\columnwidth]{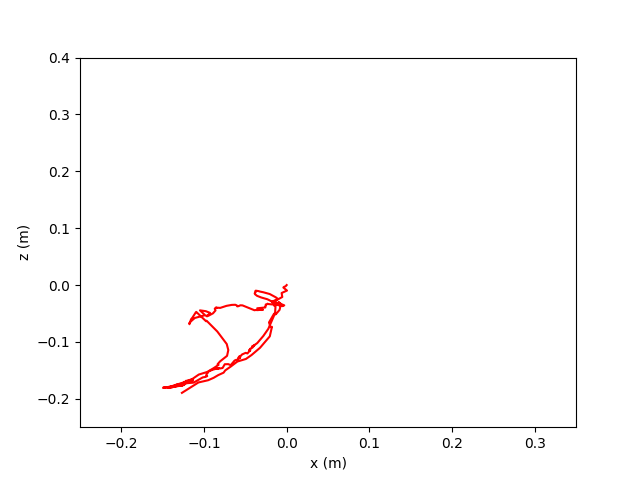}
    \end{subfigure}
    \hspace{1pt}
    \begin{subfigure}[h]{0.37\columnwidth}
        \centering
        \includegraphics[trim={10pt 5pt 40pt 35pt},clip,width=0.99\columnwidth]{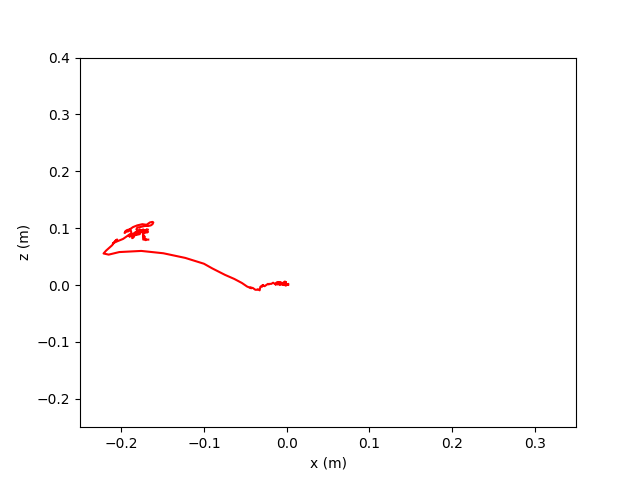}
    \end{subfigure}
    \hspace{1pt}
    \begin{subfigure}[h]{0.37\columnwidth}
        \centering
        \includegraphics[trim={10pt 5pt 40pt 35pt},clip,width=0.99\columnwidth]{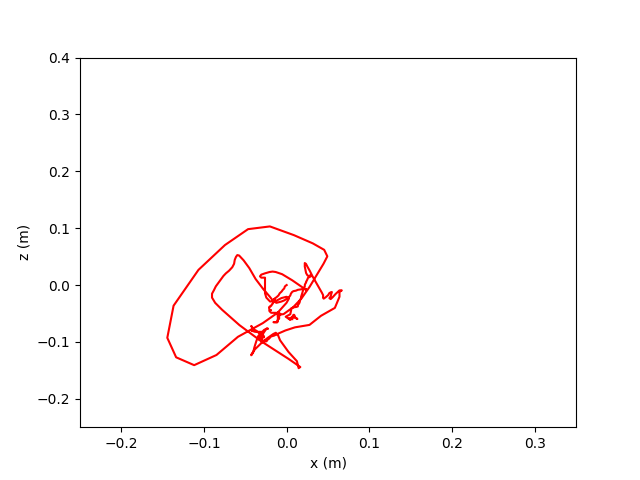}
    \end{subfigure}
    \hspace{1pt}
    \begin{subfigure}[h]{0.37\columnwidth}
        \centering
        \includegraphics[trim={10pt 5pt 40pt 35pt},clip,width=0.99\columnwidth]{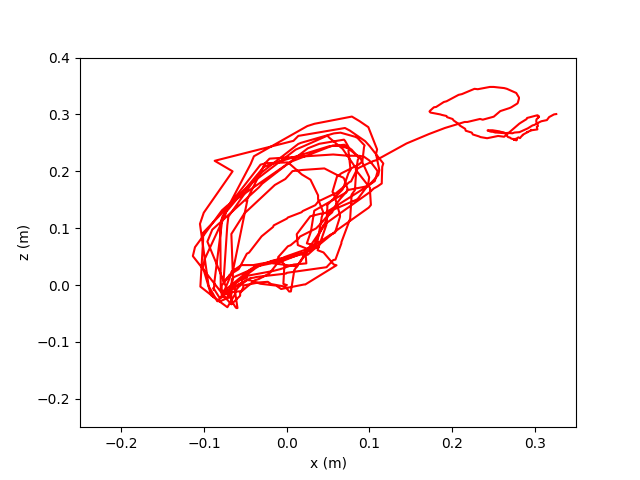}
    \end{subfigure}
    
    \vspace{3pt}
    \begin{subfigure}[h]{0.37\columnwidth}
        \centering
        \includegraphics[trim={10pt 5pt 40pt 35pt},clip,width=0.99\columnwidth]{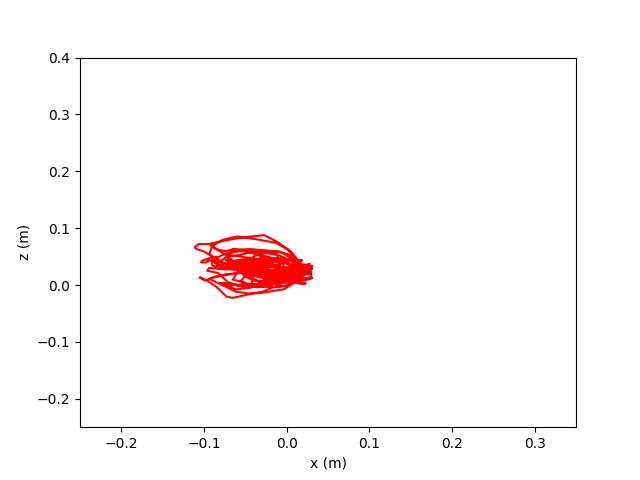}
    \end{subfigure}
    \hspace{1pt}
    \begin{subfigure}[h]{0.37\columnwidth}
        \centering
        \includegraphics[trim={10pt 5pt 40pt 35pt},clip,width=0.99\columnwidth]{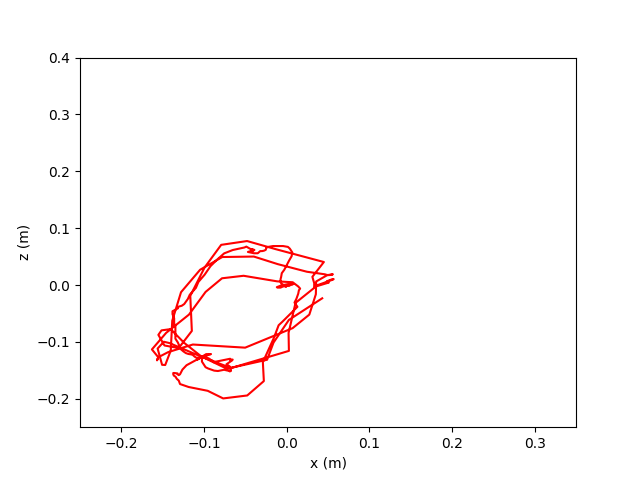}
    \end{subfigure}
    \hspace{1pt}
    \begin{subfigure}[h]{0.37\columnwidth}
        \centering
        \includegraphics[trim={10pt 5pt 40pt 35pt},clip,width=0.99\columnwidth]{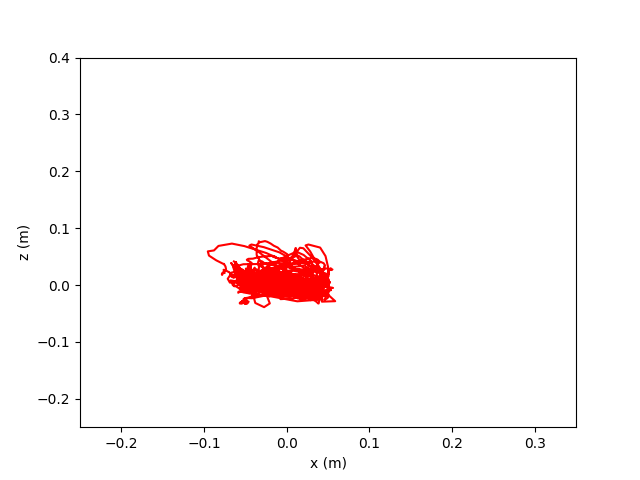}
    \end{subfigure}
    \hspace{1pt}
    \begin{subfigure}[h]{0.37\columnwidth}
        \centering
        \includegraphics[trim={10pt 5pt 40pt 35pt},clip,width=0.99\columnwidth]{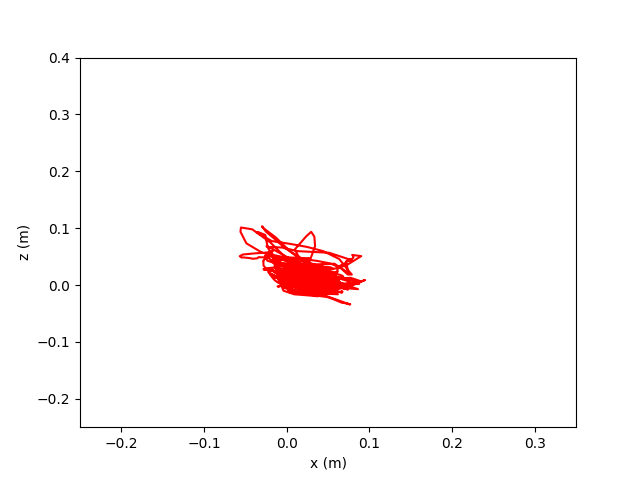}
    \end{subfigure}
    \hspace{1pt}
    \begin{subfigure}[h]{0.37\columnwidth}
        \centering
        \includegraphics[trim={10pt 5pt 40pt 35pt},clip,width=0.99\columnwidth]{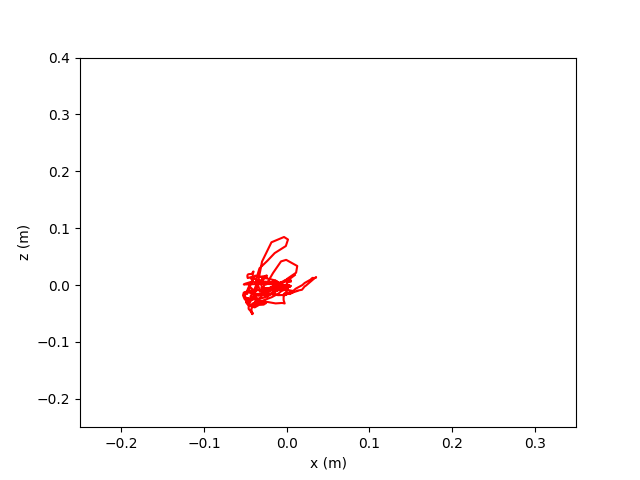}
    \end{subfigure}
\vspace{-3pt}
\caption{Sample kicking motion paths from each subject. Panels corresponding to S1 through S10 are shown from left to right and from top to bottom. \label{fig:2cell} \label{xy2}}
\vspace{-6pt}
\end{figure*}

\begin{table*}[t]
\vspace{6pt}
    \centering
    \caption{Mean Squared Error (All Values in $\times10^{-3}$)}
\label{tab:MSE}
\vspace{-3pt}
\scalebox{1}{
\begin{tabular}{lccccccccccccc}
\toprule
&  S1-T1 &  S2-T1 &  S2-T4 &  S2-T5 &  S3-T3 &  S4-T1 &  S5-T2 &  S6-T1 &  S7-T2 &  S8-T1 &  S9-T1 &  S9-T2 &  S10-T1 \\
\midrule
x Sim &   0.039 &   0.024 &   0.014 &   0.027 &   0.017 &   0.046 &   0.770 &   0.085 &   0.161 &   0.017 &   0.064 &   0.054 &   0.010 \\
x PID &   2.192 &   0.704 &   0.535 &   0.927 &   1.252 &   0.994 &   6.332 &   1.372 &   4.969 &   0.750 &   0.943 &   1.126 &   0.619 \\
x Non Lin &   1.834 &   0.716 &   0.583 &   0.774 &   0.740 &   1.063 &   4.974 &   1.507 &   4.642 &   0.576 &   1.027 &   1.121 &   0.553 \\
z Sim &   0.032 &   0.020 &   0.016 &   0.015 &   0.006 &   0.072 &   1.599 &   0.024 &   0.218 &   0.006 &   0.021 &   0.034 &   0.024 \\
z PID &   1.714 &   2.021 &   2.544 &   1.722 &   1.536 &   2.413 &  10.709 &   0.759 &   5.073 &   0.341 &   0.548 &   1.969 &   1.654 \\
z Non Lin &   0.625 &   0.415 &   0.765 &   0.328 &   0.213 &   1.098 &   5.896 &   0.289 &   2.286 &   0.158 &   0.218 &   0.391 &   0.297 \\
\bottomrule
\end{tabular}}
\end{table*}


\begin{table*}[t]
\vspace{-3pt}
    \centering
    \caption{Dynamic Time Warping Minimum Path Distance}
\label{tab:minPath}
\vspace{-3pt}
\scalebox{1}
{\begin{tabular}{lccccccccccccc}
\toprule
&        S1-T1 &  S2-T1 &  S2-T4 &  S2-T5 &  S3-T3 &  S4-T1 &  S5-T2 &  S6-T1 &  S7-T2 &  S8-T1 &        S9-T1 &   S9-T2 &    S10-T1 \\
\midrule
x Sim &  1.447 &  0.504 &  0.341 &  0.490 &   0.234 &  0.578 &  4.153 &  2.463 &  0.852 &   3.773 &  2.515 &  0.511 &  0.483 \\
x PID &  4.803 &  1.809 &  1.757 &  2.664 &   2.505 &  2.953 &  9.875 & 12.885 &  4.412 &  23.057& 16.733 &  3.524 & 4.577 \\
x Non Lin &  5.225 &  1.789 &  1.853 & 2.232 &    2.110 &  2.155 &  7.989 &  10.088 &  3.549 &  21.652 &  15.170 &  2.906 &  5.197 \\
z Sim &  0.899 &  0.428 &  0.319 &  0.301 &   0.141 &  0.767 &  5.733 &  1.044 &  0.936 &   2.575 &  1.314 &  0.330 &  0.524 \\
z PID &  7.304 &  4.418 &  3.378 &  5.085 &   3.918 &  6.751 & 13.031 &  6.979 &  4.666 & 28.959 &  11.788 &  3.571 &  8.375 \\
z Non Lin &  3.615 &   1.736 &   1.848 &   1.896 & 1.462 &  2.256 &  7.868 &  4.815 &  2.193 &  21.810 & 6.461 &  1.401 &  3.154 \\
\bottomrule
\end{tabular}}
\vspace{-12pt}
\end{table*}


While dynamics~\eqref{eq:translationalDyn}--\eqref{eq:rotationalDyn} with control input~\eqref{eq:input} are necessary to describe the robot's motion, they are not sufficient to link to specific tasks for the robot, such as for example waypoint navigation or trajectory tracking (as in this work). In such cases it is needed to add a higher-level controller that tracks the evolution of robot state (or a subset of it, for instance positions in space) over time and determines lower-level control inputs~\eqref{eq:input} to minimize state tracking errors. 
To this end, we considered position control of the aerial robot. Desired position commands matched the extracted 2D tracked position of infant kicking trajectories, and were sent to the robot at rates matching the input video frame rates (Table~\ref{tab:subjects}). While other types of controllers are applicable (e.g., velocity control or commanding accelerations directly), we elected here to use position control so as to match the type of input data and measured data available to us (in both cases positions).\footnote{~While in simulation we have access to simulated robot velocity, in physical experiments we only had access to measured robot position. To ensure the gap between simulation and physical experiments remains as small as possible, we used simulated robot position alone in simulation.} We have considered two types of controllers, a proportional-integral-derivative (PID) controller tested in physical experiments, as well as a nonlinear controller~\cite{lee2010geometric,mellinger2011minimum} tested in both simulation and physical experiments. Specific details for the simulation and physical experiments are provided next.

 


\subsection{Simulation and Physical Experiments Setup}
Simulations were run using the Robot Operating System (ROS). Main files to input desired trajectories, collect and log the robot's output position, and to analyze collected position data were written in Python. For execution of robot commands we used an open-source package~\cite{preiss2017crazyswarm} integrated with ROS. The package provides an implementation of the nonlinear controller employed herein~\cite{lee2010geometric,mellinger2011minimum}. For state propagation in simulation, computed lower-level control inputs were passed on dynamics~\eqref{eq:translationalDyn}--\eqref{eq:rotationalDyn} and output thrust and moments were used to determine the next state of the simulated robot based on rigid body motion fundamentals. The updated state was then fed back to higher-level controller, and the loop continues.


%
The physical experiments follow closely the simulation setup. 
The nonlinear controller is exactly the same as in simulation. For the linear PID controller we used a cascaded control structure. Only the position is directly controlled by the algorithm, whereas the robot's attitude is determined implicitly by the position controller and then fed into an attitude controller to determine desired angular rates. Computed linear accelerations and angular rates are then passed via the robot's equations of motion to link to desired motor thrusts. The robot's firmware (provided by the manufacturer) maps commanded thrusts into rotor angular velocities. 
A 32-bit, $168$\;MHz ARM microcontroller is responsible for onboard flight control computation. Desired commands (e.g., waypoints) are transmitted from a PC workstation to the robot wirelessly via a $2.4$\;GHz USB radio. As in simulation, the same open-source software~\cite{preiss2017crazyswarm} was used to communicate high-level commands to the robot, while the robot's firmware ensures the computed rotor angular velocities are achieved. All software was run in ROS, which facilitates code re-use between simulation and physical experiments. Robot real-time position and orientation measurements were provided by an 8-camera motion capture system (Optitrack Prime X22). 
 





\begin{figure}[!t]
\vspace{4pt}
\centering
    \begin{subfigure}[h]{0.49\columnwidth}
        \centering
        \includegraphics[trim={10pt 0pt 10pt 40pt},clip,width=1.05\columnwidth]{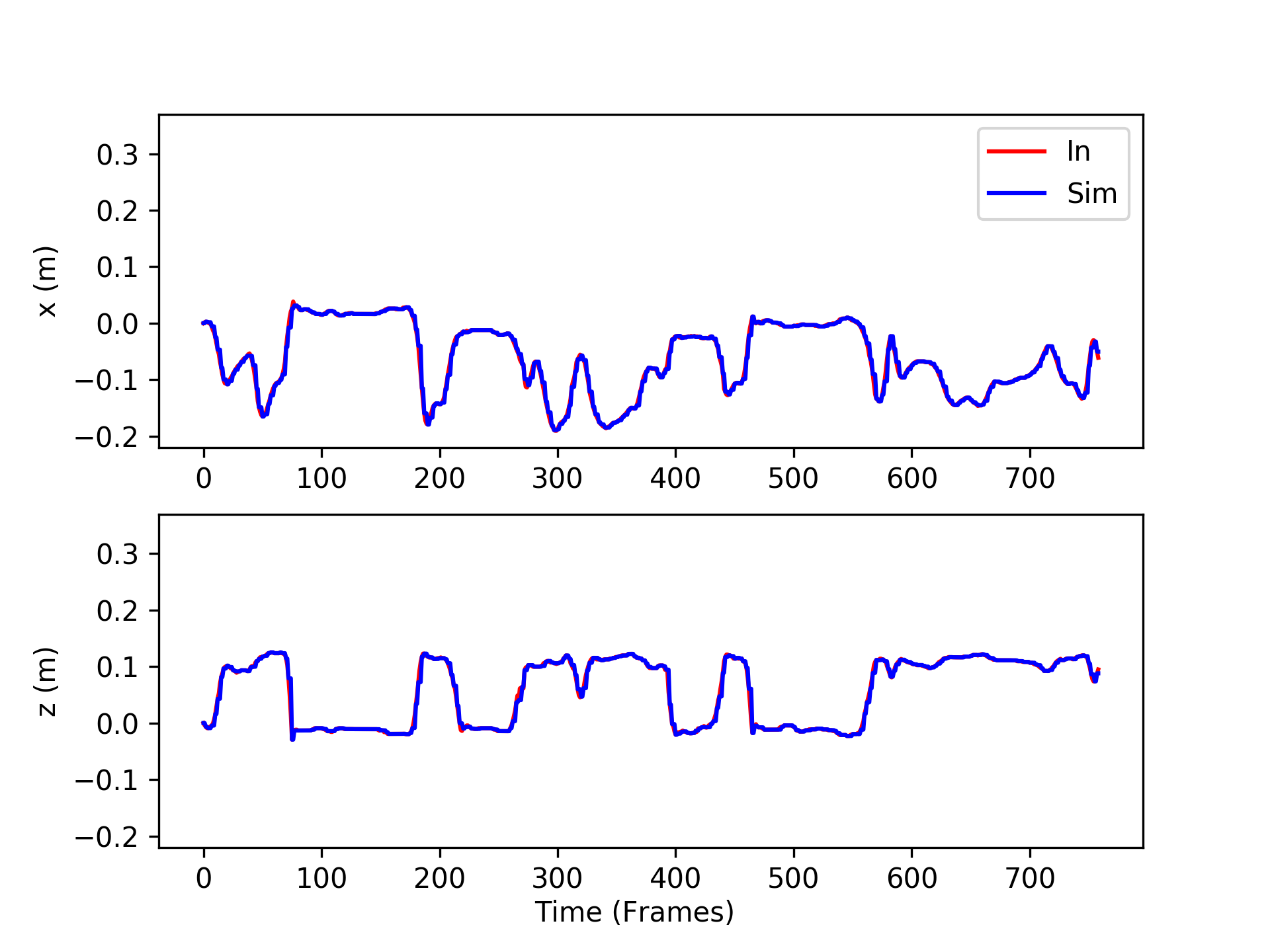}
    \end{subfigure}
    \hfill
    \begin{subfigure}[h]{0.49\columnwidth}
        \centering
        \includegraphics[trim={10pt 0pt 10pt 40pt},clip,width=1.05\columnwidth]{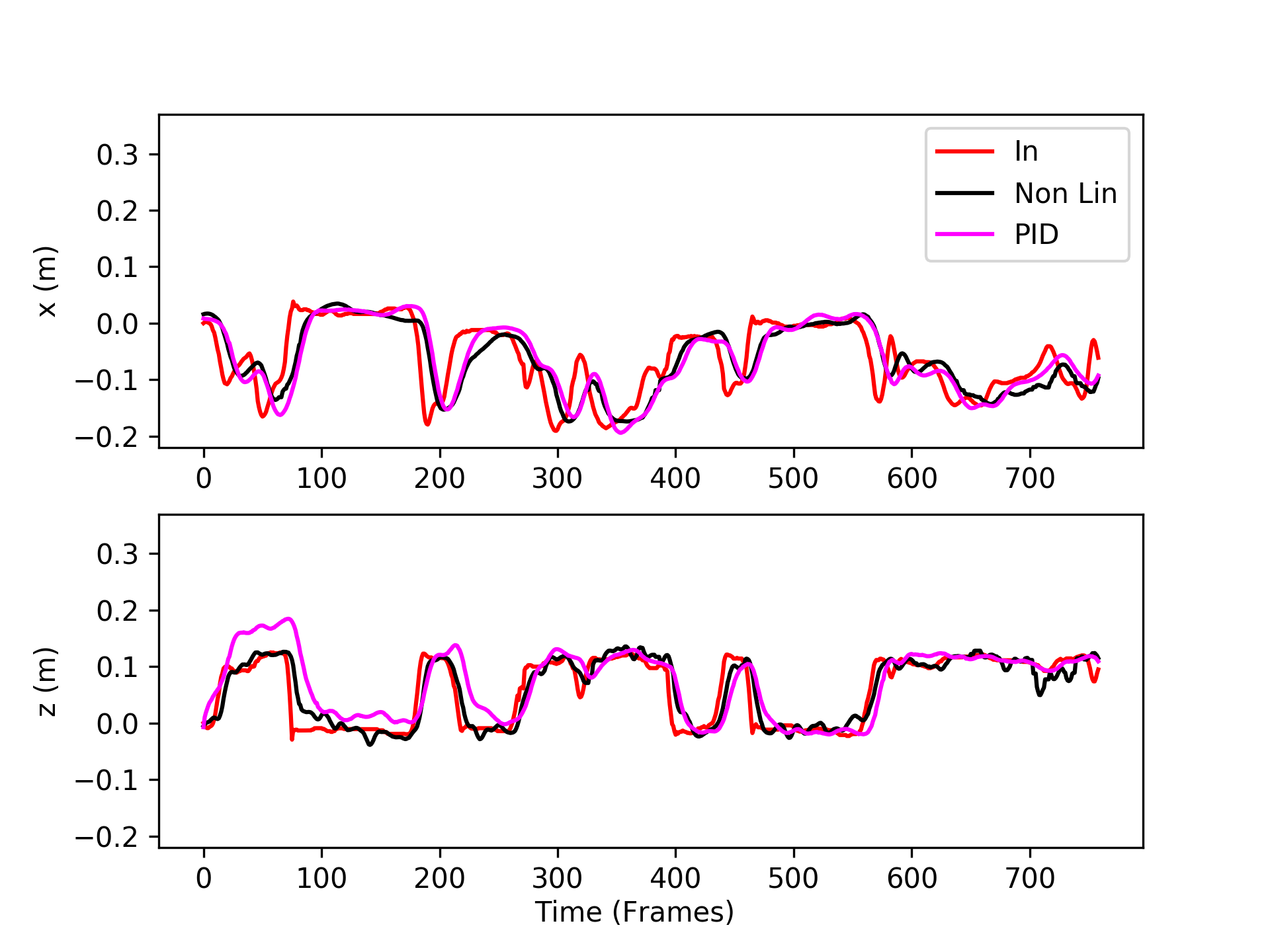}
    \end{subfigure}
    
    \vspace{3pt}
    \begin{subfigure}[h]{0.49\columnwidth}
        \centering
        \includegraphics[trim={10pt 0pt 10pt 40pt},clip,width=1.05\columnwidth]{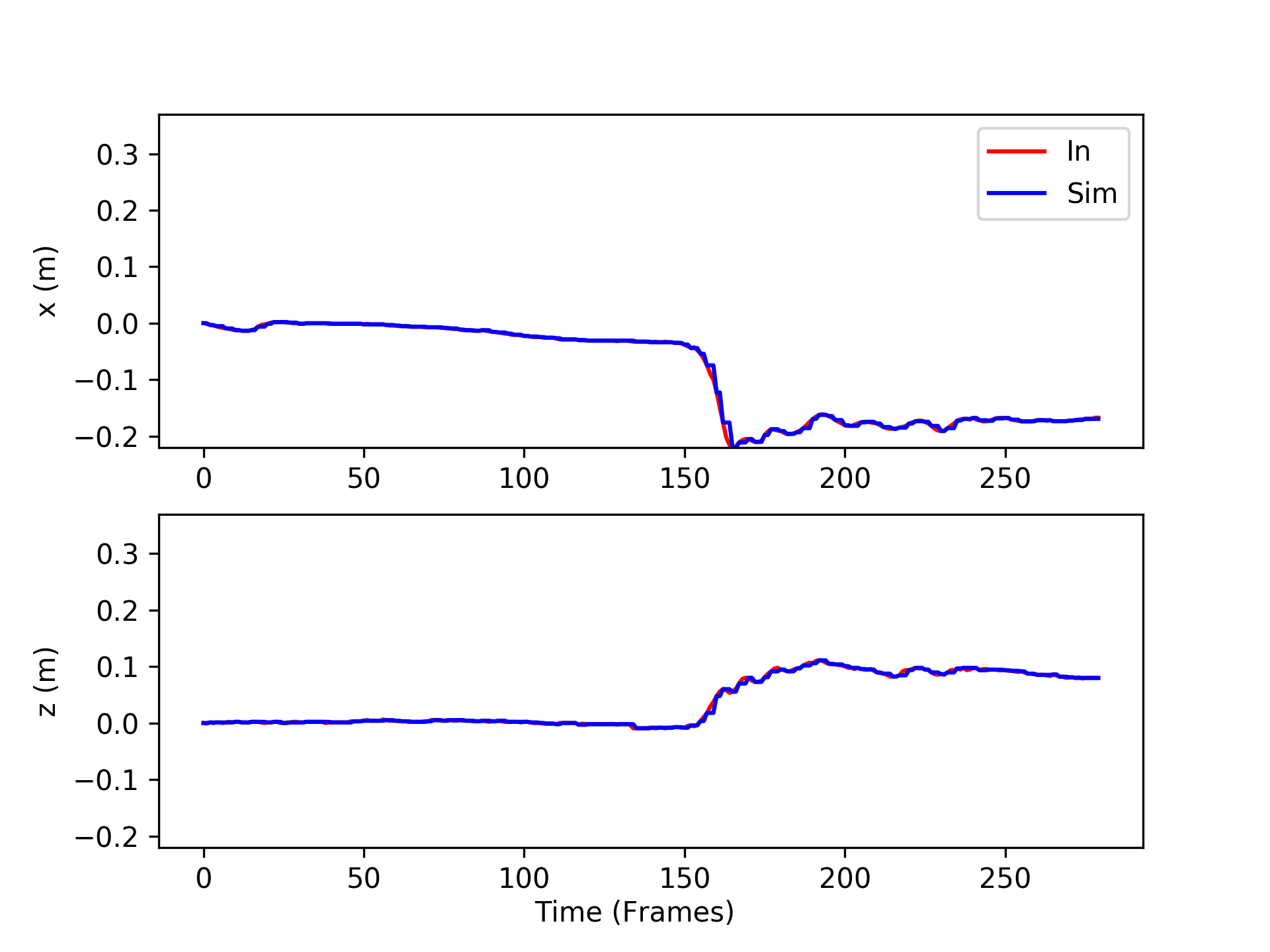}
    \end{subfigure}
    \hfill
    \begin{subfigure}[h]{0.49\columnwidth}
        \centering
        \includegraphics[trim={10pt 0pt 10pt 40pt},clip,width=1.05\columnwidth]{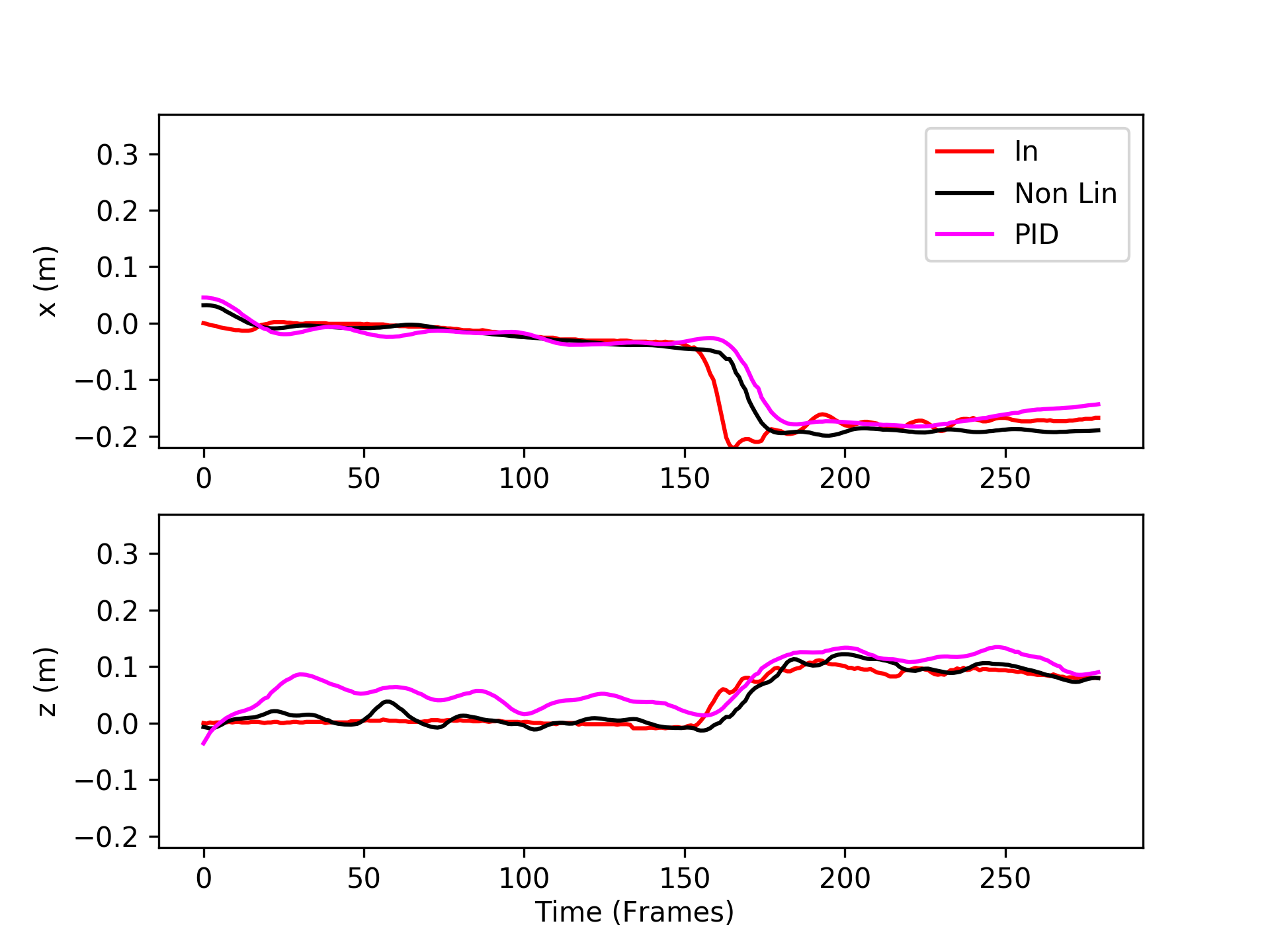}
    \end{subfigure}
    
    \vspace{3pt}
    \begin{subfigure}[h]{0.49\columnwidth}
        \centering
        \includegraphics[trim={10pt 0pt 10pt 40pt},clip,width=1.05\columnwidth]{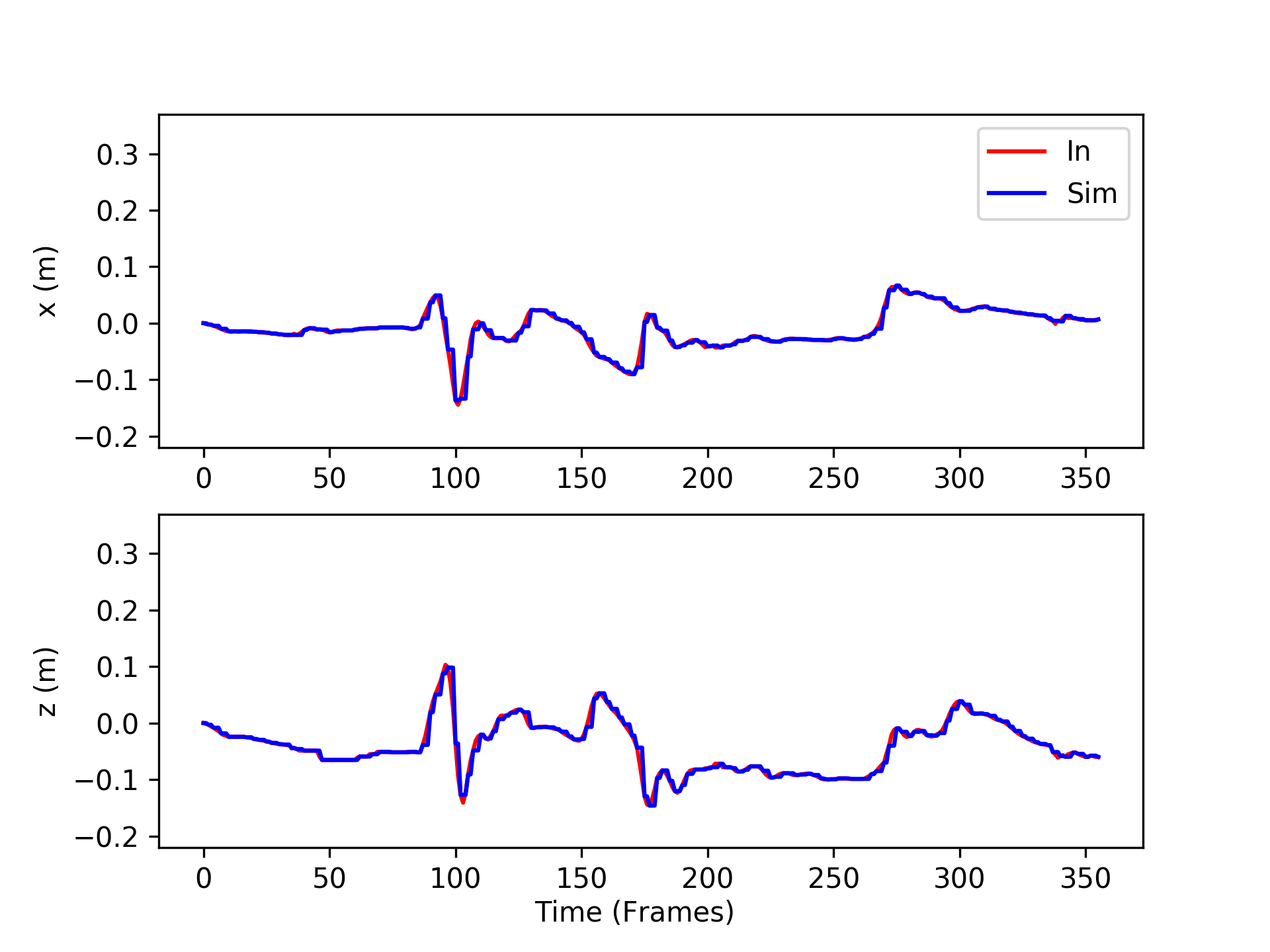}
    \end{subfigure}
    \hfill
    \begin{subfigure}[h]{0.49\columnwidth}
        \centering
        \includegraphics[trim={10pt 0pt 10pt 40pt},clip,width=1.05\columnwidth]{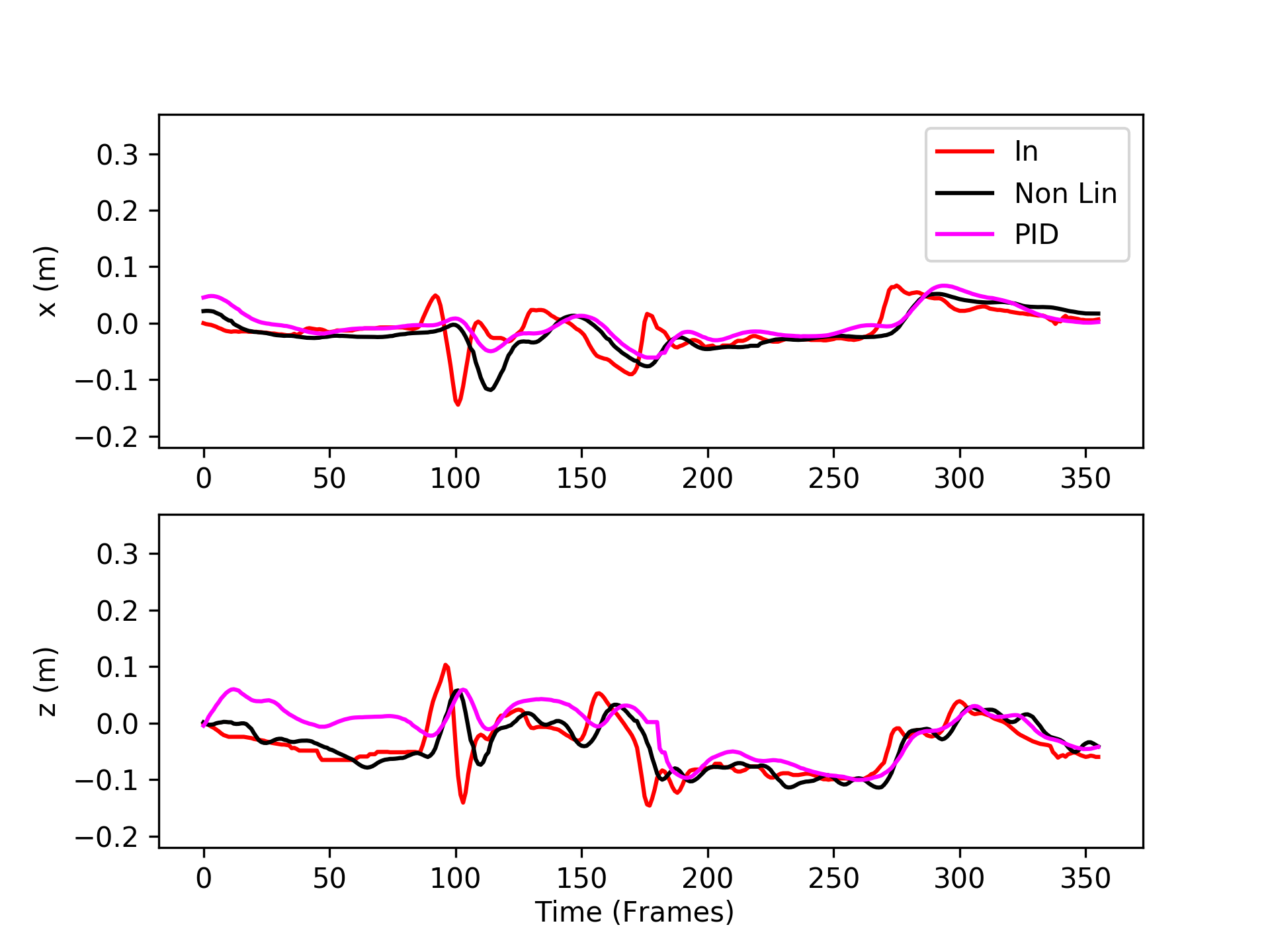}
    \end{subfigure}
    
    \vspace{3pt}
    \begin{subfigure}[h]{0.49\columnwidth}
        \centering
        \includegraphics[trim={10pt 0pt 10pt 40pt},clip,width=1.05\columnwidth]{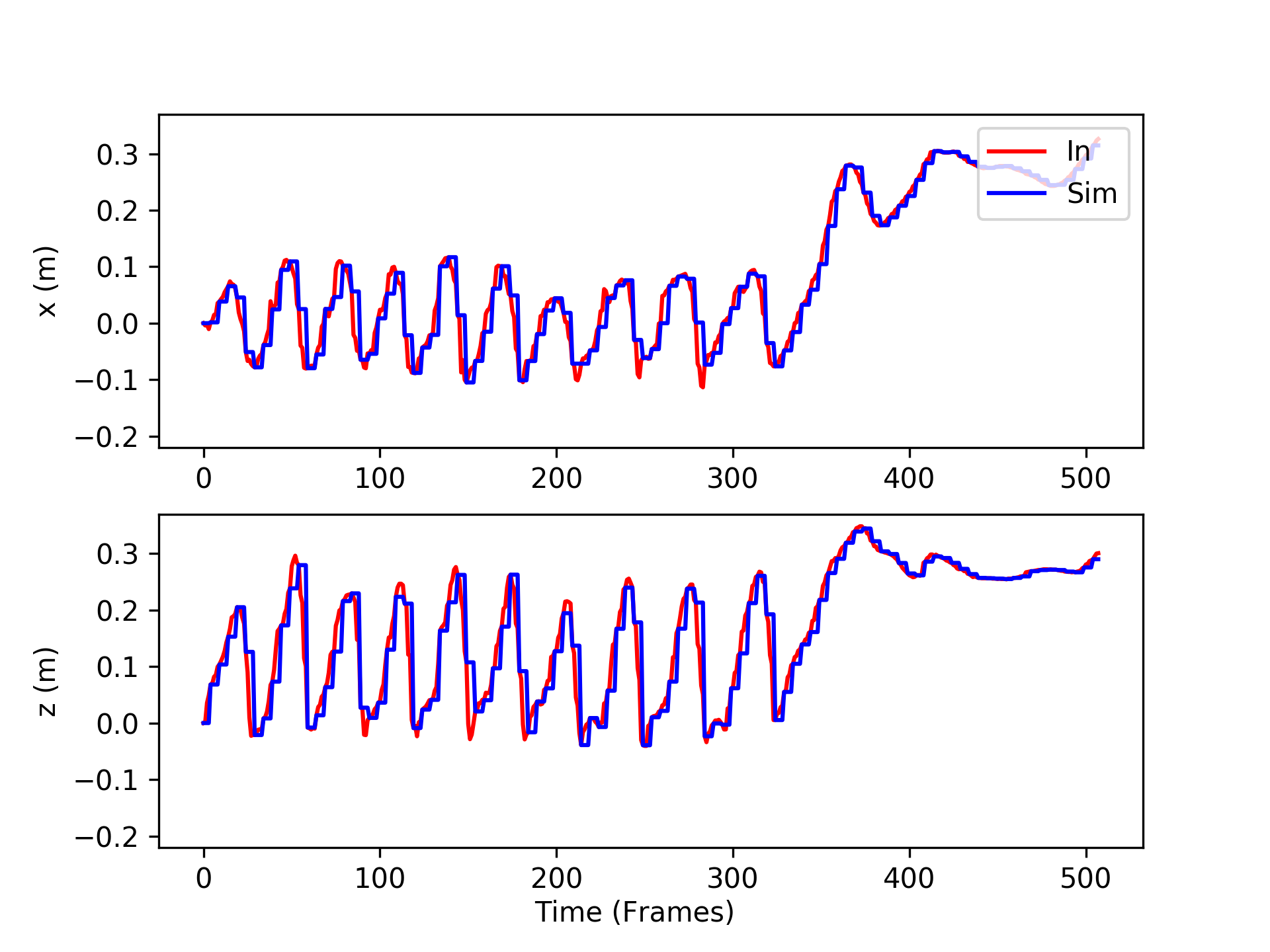}
    \end{subfigure}
    \hfill
    \begin{subfigure}[h]{0.49\columnwidth}
        \centering
        \includegraphics[trim={10pt 0pt 10pt 40pt},clip,width=1.05\columnwidth]{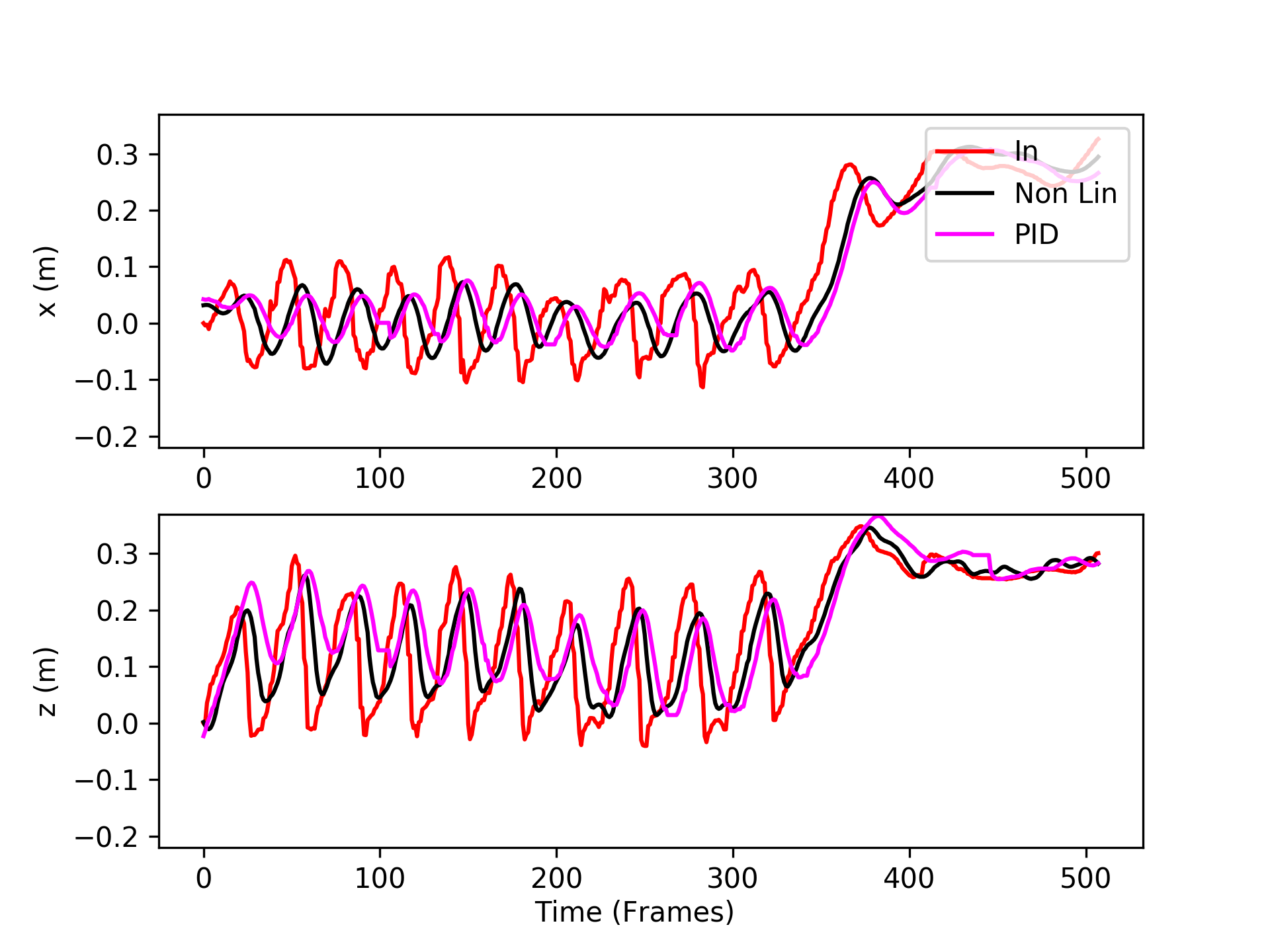}
    \end{subfigure}
    
    \vspace{3pt}
    \begin{subfigure}[h]{0.49\columnwidth}
        \centering
        \includegraphics[trim={10pt 0pt 10pt 40pt},clip,width=1.05\columnwidth]{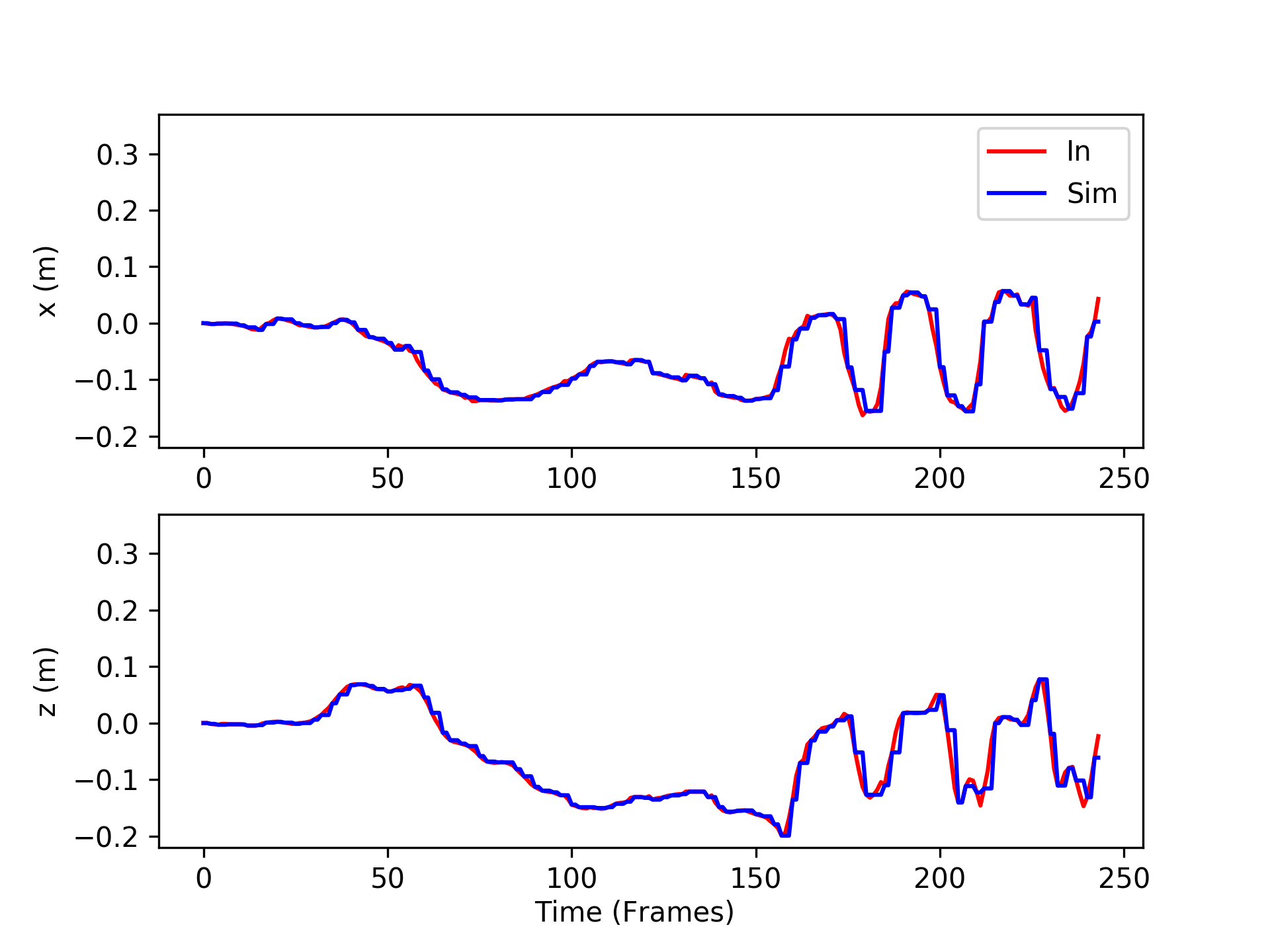}
    \end{subfigure}
    \hfill
    \begin{subfigure}[h]{0.49\columnwidth}
        \centering
        \includegraphics[trim={10pt 0pt 10pt 40pt},clip,width=1.05\columnwidth]{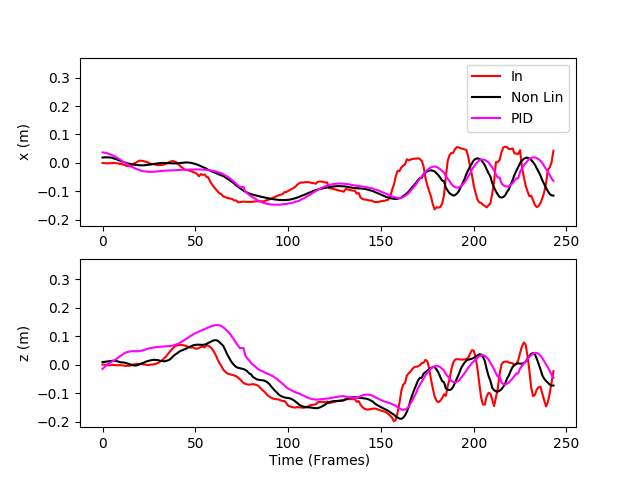}
    \end{subfigure}
    
    \vspace{3pt}
    \begin{subfigure}[h]{0.49\columnwidth}
        \centering
        \includegraphics[trim={10pt 0pt 10pt 40pt},clip,width=1.05\columnwidth]{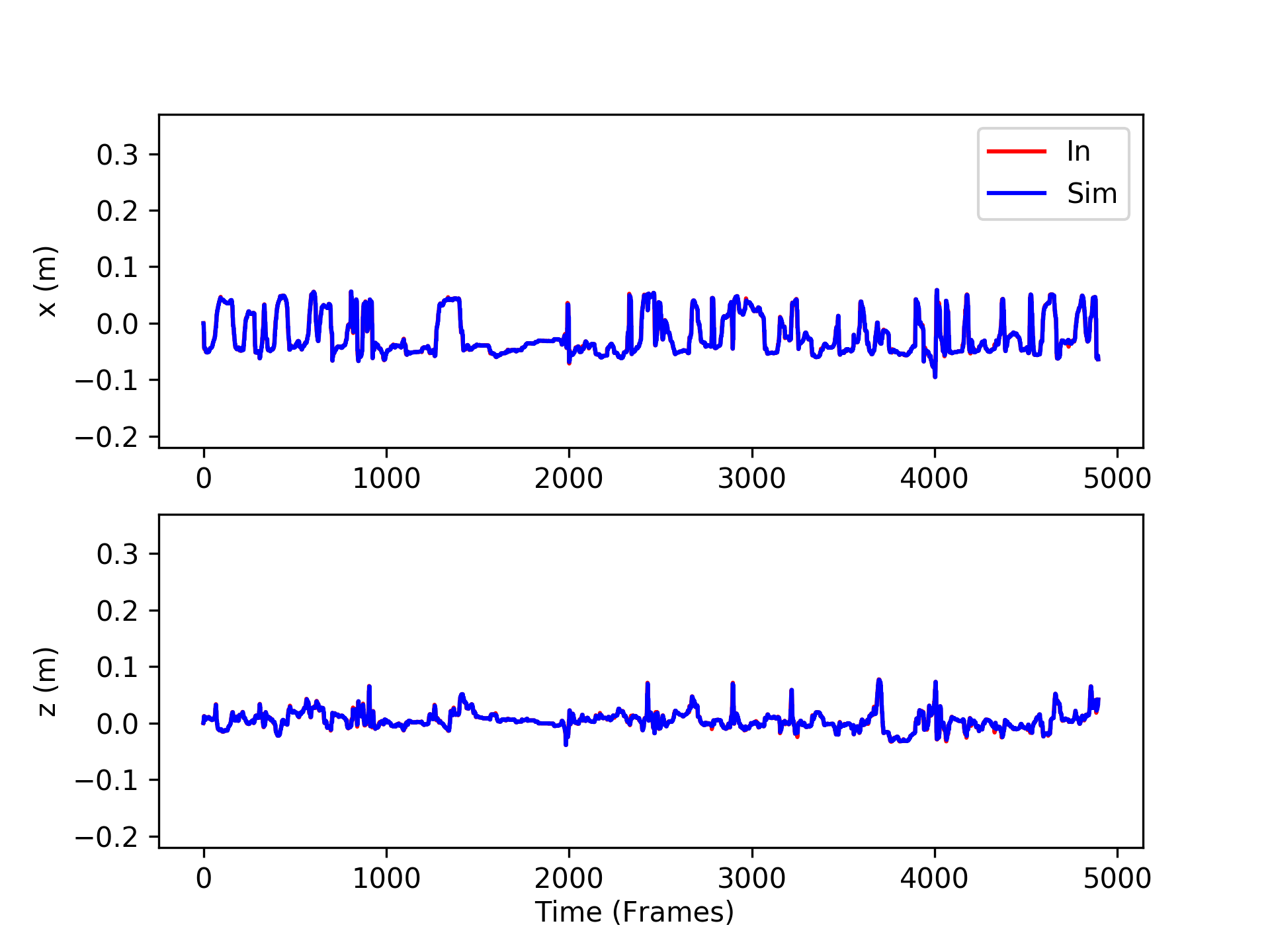}
    \end{subfigure}
    \hfill
    \begin{subfigure}[h]{0.49\columnwidth}
        \centering
        \includegraphics[trim={10pt 0pt 10pt 40pt},clip,width=1.05\columnwidth]{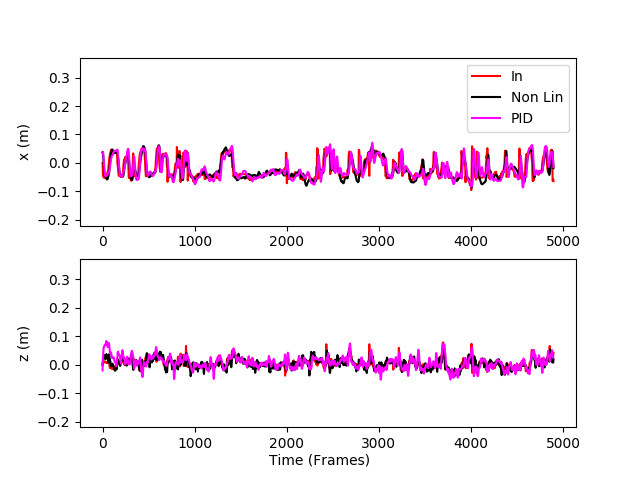}
    \end{subfigure}
    \vspace{-3pt}
\caption{Examples of leg position data (In) obtained from different subjects and position data obtained from the simulation (left) and physical experiments (right) with the quadrotor (top to bottom, subjects: S1, S3, S4, S5, S7, S8). \label{fig3}}
\vspace{-20pt}
\end{figure}


\section{Results and Discussion}
Overall, simulation results demonstrate that the quadrotor can track infant leg motion trajectories across all cases. Tracking infant trajectories in physical experiments with the robot is also attainable, yielding in the meantime interesting insights regarding the various types of motion that can be tracked by the current standard of practice quadrotor controllers (linear PID and nonlinear geometric). 

Figure~\ref{fig3} highlights leg position trajectories and outputs from simulation and physical experiments in the most illustrative case studies. From a qualitative analysis standpoint, we observe that in simulation the robot can track well the leg trajectory when the latter is aperiodic.
Tracking periodic signals (panels in rows 4 and 6 in Fig.~\ref{fig3}) is found to be less efficient as it is possible to have phase lag and amplitude errors.
We also observe that tracking of very fast increasing in amplitude signals (panels in rows 2 and 3 in Fig.~\ref{fig3}) is possible but with some delay.
Qualitative analysis of simulation results indicates that tracking infant kicking motion with an aerial robot is in principle feasible. In physical experiments, the same observations made in simulation also hold. However, we notice much more pronounced phase lag (in periodic parts of the signals) and delays (in fast increasing in amplitude signals).
We also notice that some times rapid movements with opposite velocity (i.e. chattering portions of the signal) are filtered out by the controller which makes the robot track the average of the signal (best observed at panels in row 3 in Fig.~\ref{fig3}). Further, the nonlinear controller has a better performance overall than the linear PID controller, especially as the input signal becomes more variable.

These qualitative findings are also supported via quantitative analysis. 
First, we computed the mean-squared error (MSE) between ground-truth data and the quadrotor output positions for each case (and both in simulation and physical experiments). The MSE can provide a quantitative score that describes the level of error using an one-to-one correspondence between the elements of the two time series being compared. With reference to Table~\ref{tab:MSE}, in simulation the MSE attains lower values compared to physical experiments, consistent with the qualitative analysis. As discussed above, physical limitations and system delays when using the physical quadrotor contribute to higher MSE. Minimum MSE values appear without any specific pattern in different tracks for both controllers that were tested (PID, and nonlinear). The maximum MSE was observed in S5-T2 where the infant's motion rapidly forms a quick sustained cyclic pattern but the robot is unable to reach the same magnitude at the same rate as the infant's sustained kicking pattern. For S4-T1 there are no significant variations in motion along the x axis as the infant maintains an approximately $90$\;deg knee flexion, which in turn results in a stable MSE for both controllers. Overall, all MSE values computed are less than $0.01$\;m$^2$.

\begin{figure}[!h]
\vspace{6pt}
\centering
\includegraphics[trim={40pt 50pt 40pt 85pt},clip,width=0.85\columnwidth]{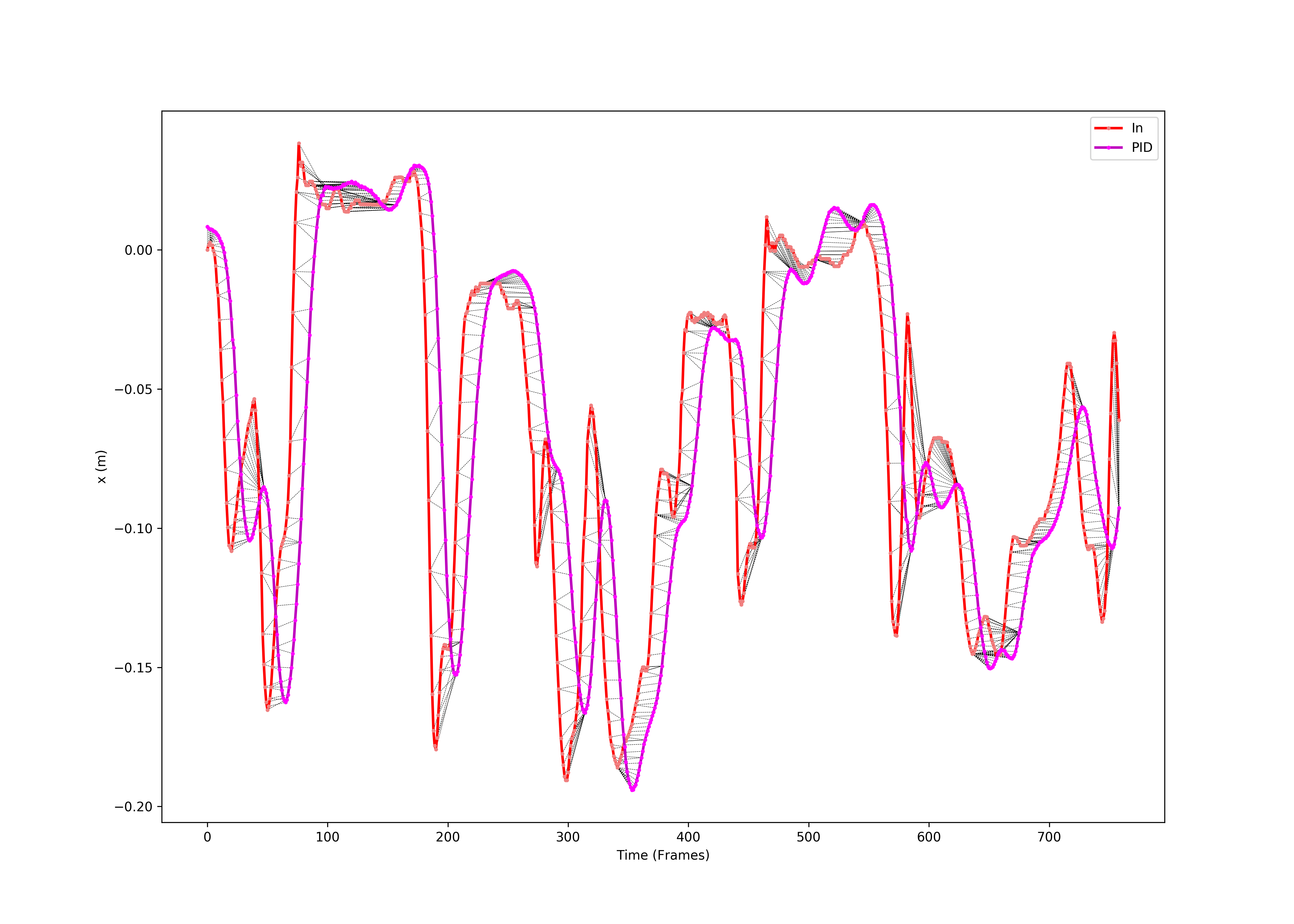}
\vspace{0pt}
\caption{Sample Dynamic Time Warping algorithm matching of x-axis components between input time series and output of the PID controller in physical experiments for S1-T1. \label{fig5}}
\end{figure}

\begin{figure}[!h]
\vspace{-12pt}
\centering
\includegraphics[trim={40pt 50pt 40pt 85pt},clip,width=0.85\columnwidth]{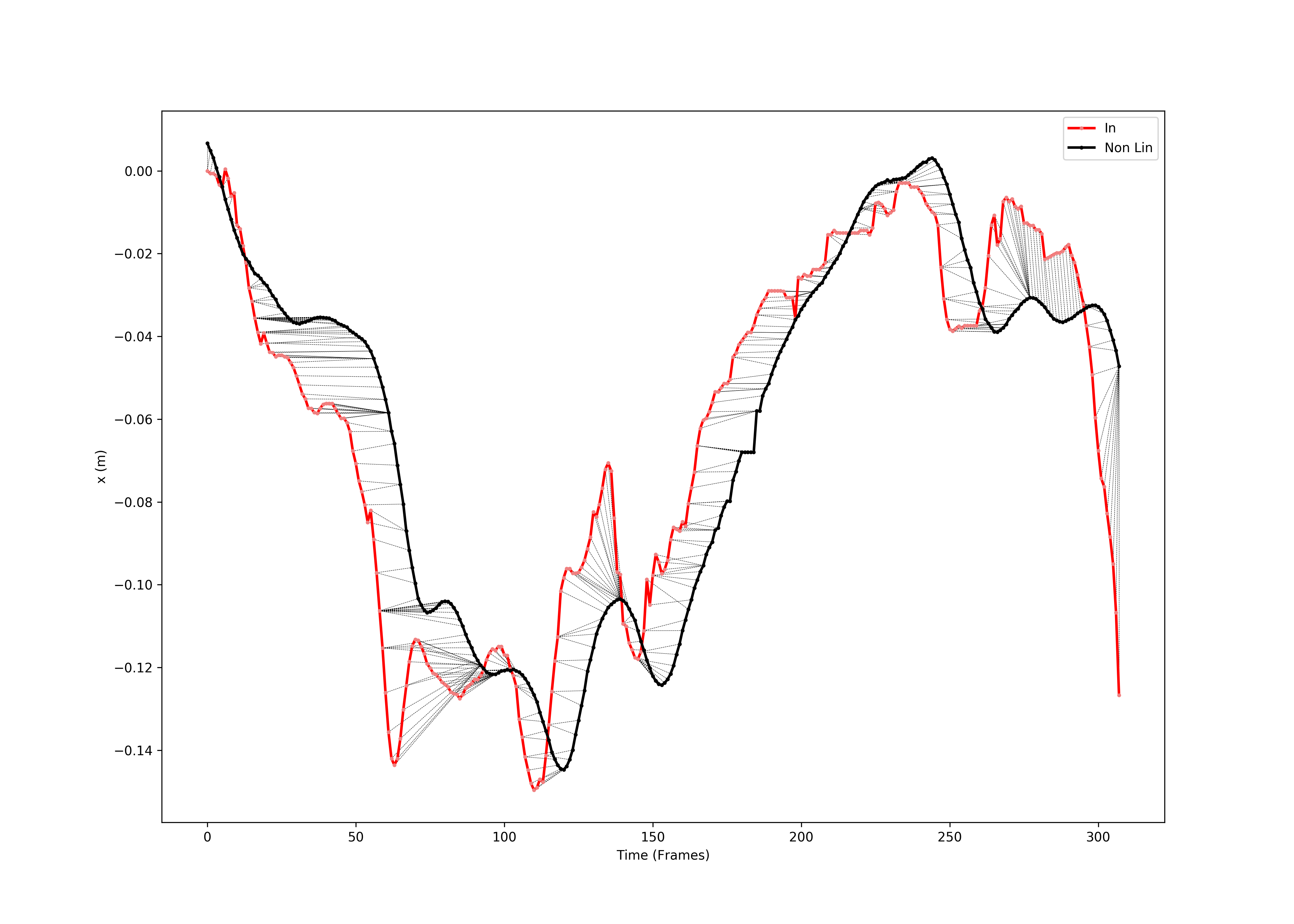}
\vspace{0pt}
\caption{Sample Dynamic Time Warping algorithm matching of x-axis components between input time series and output of the nonlinear controller in physical experiments for S2-T4.\label{fig6}}
\vspace{-6pt}
\end{figure}

\begin{figure}[!t]
\vspace{6pt}
\centering
    \begin{subfigure}[h]{0.33\columnwidth}
        \centering
        \includegraphics[trim={52pt 0pt 40pt 40pt},clip,width=1.15\columnwidth]{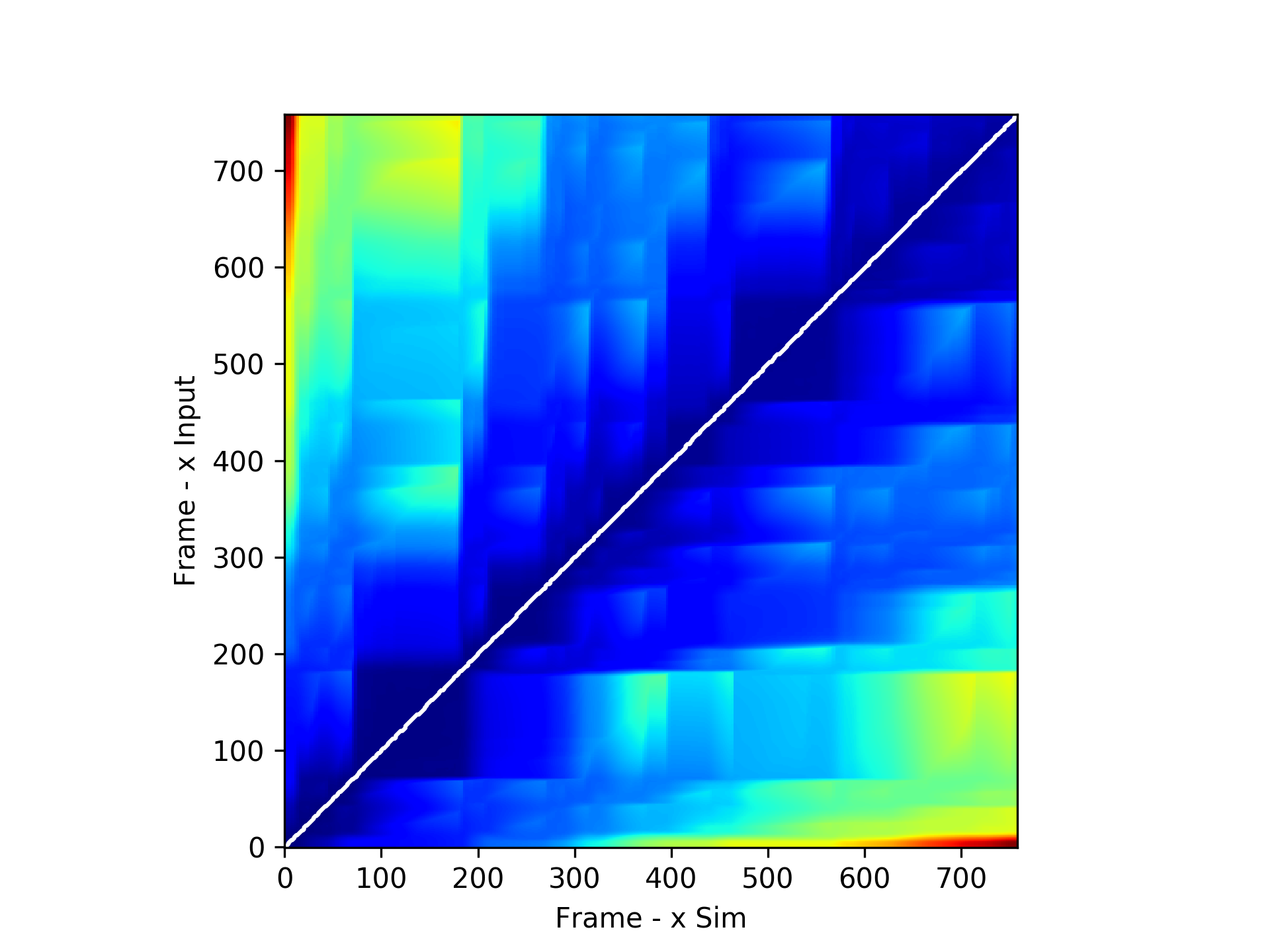}
    \end{subfigure}
    \hspace{-6pt}
    \begin{subfigure}[h]{0.33\columnwidth}
        \centering
        \includegraphics[trim={52pt 0pt 40pt 40pt},clip,width=1.15\columnwidth]{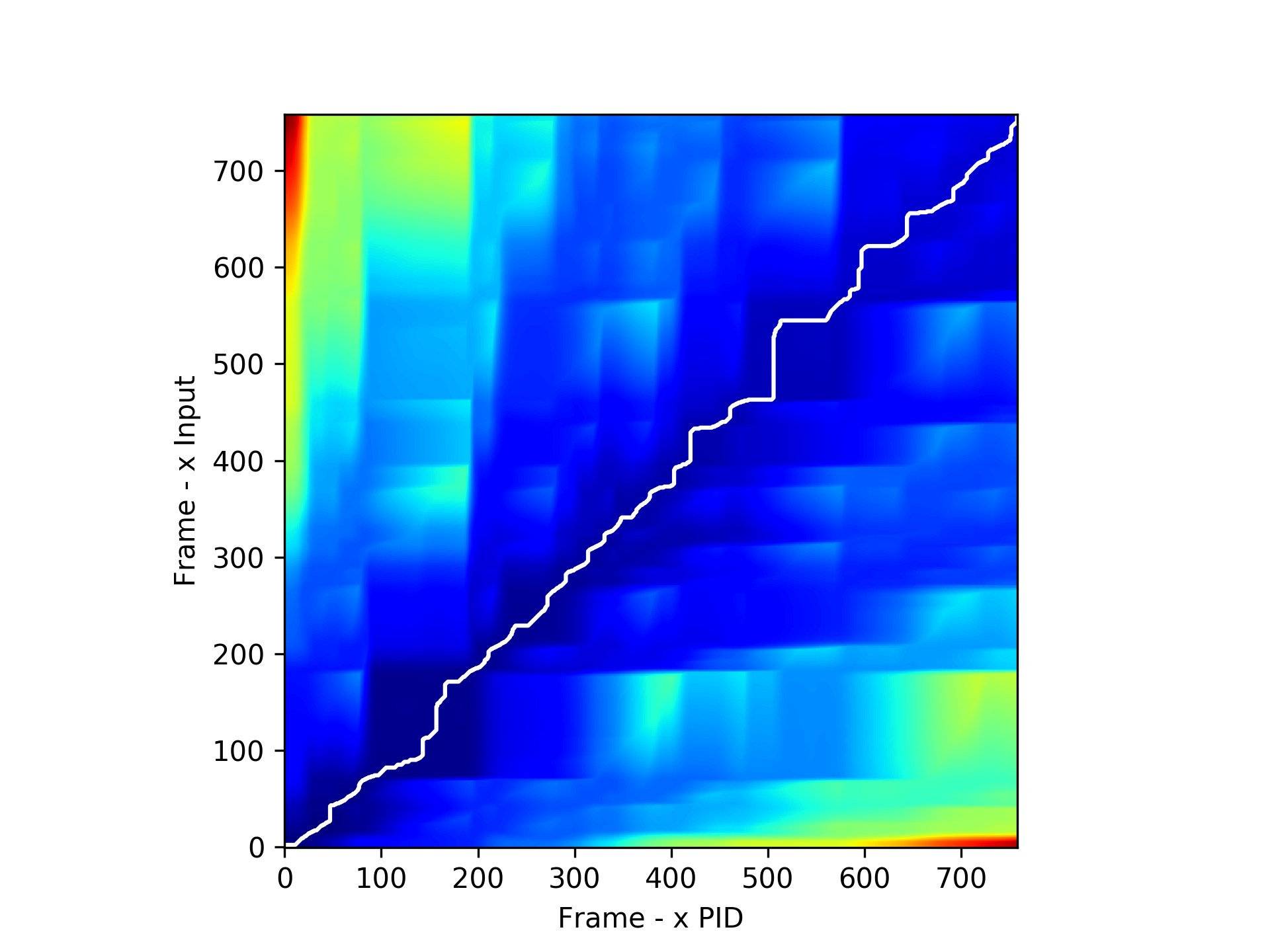}
    \end{subfigure}
    \hspace{-6pt}
    \begin{subfigure}[h]{0.33\columnwidth}
        \centering
        \includegraphics[trim={52pt 0pt 40pt 40pt},clip,width=1.15\columnwidth]{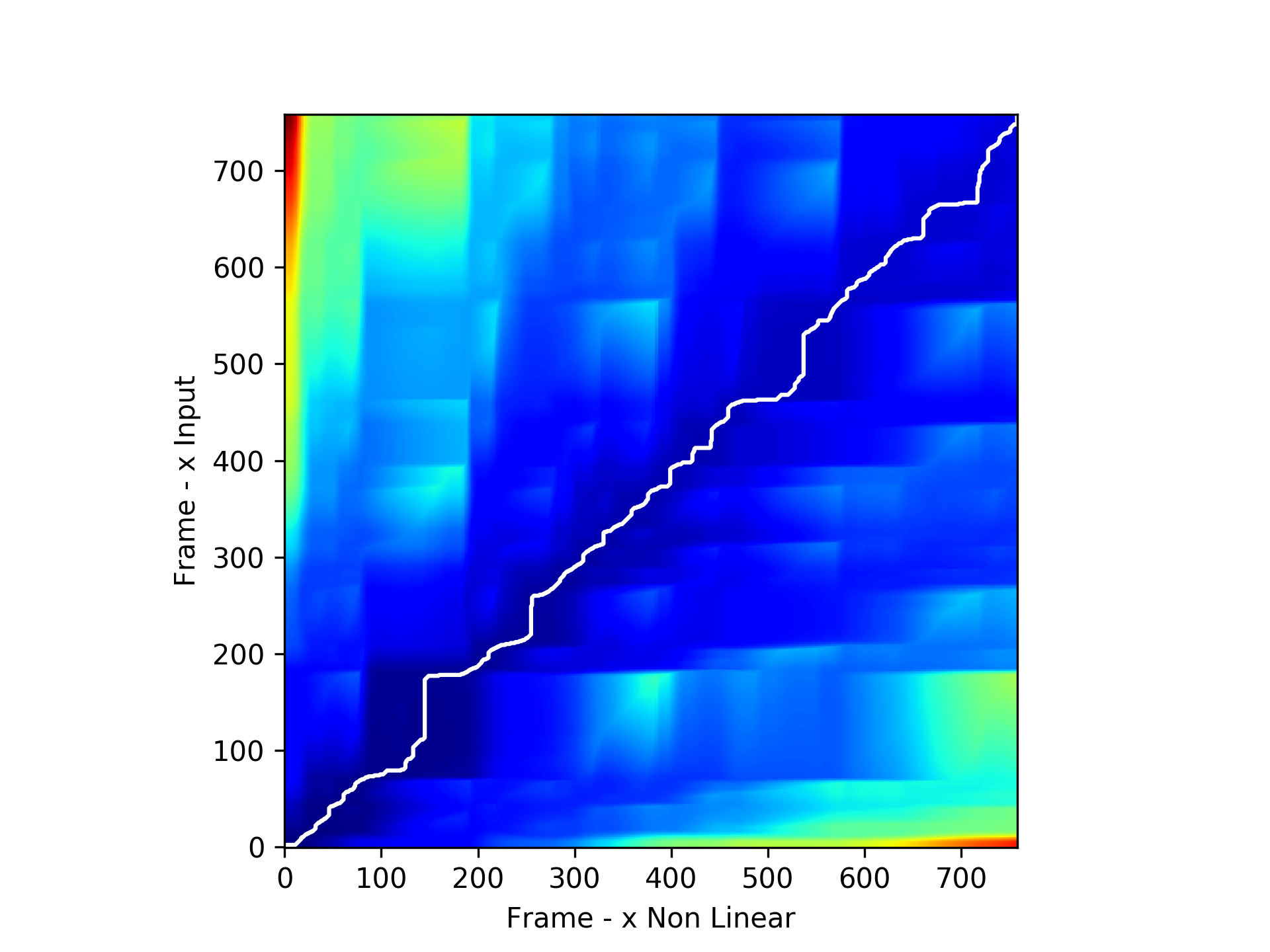}
    \end{subfigure}

    \vspace{3pt}
    \begin{subfigure}[h]{0.33\columnwidth}
        \centering
        \includegraphics[trim={52pt 0pt 40pt 40pt},clip,width=1.15\columnwidth]{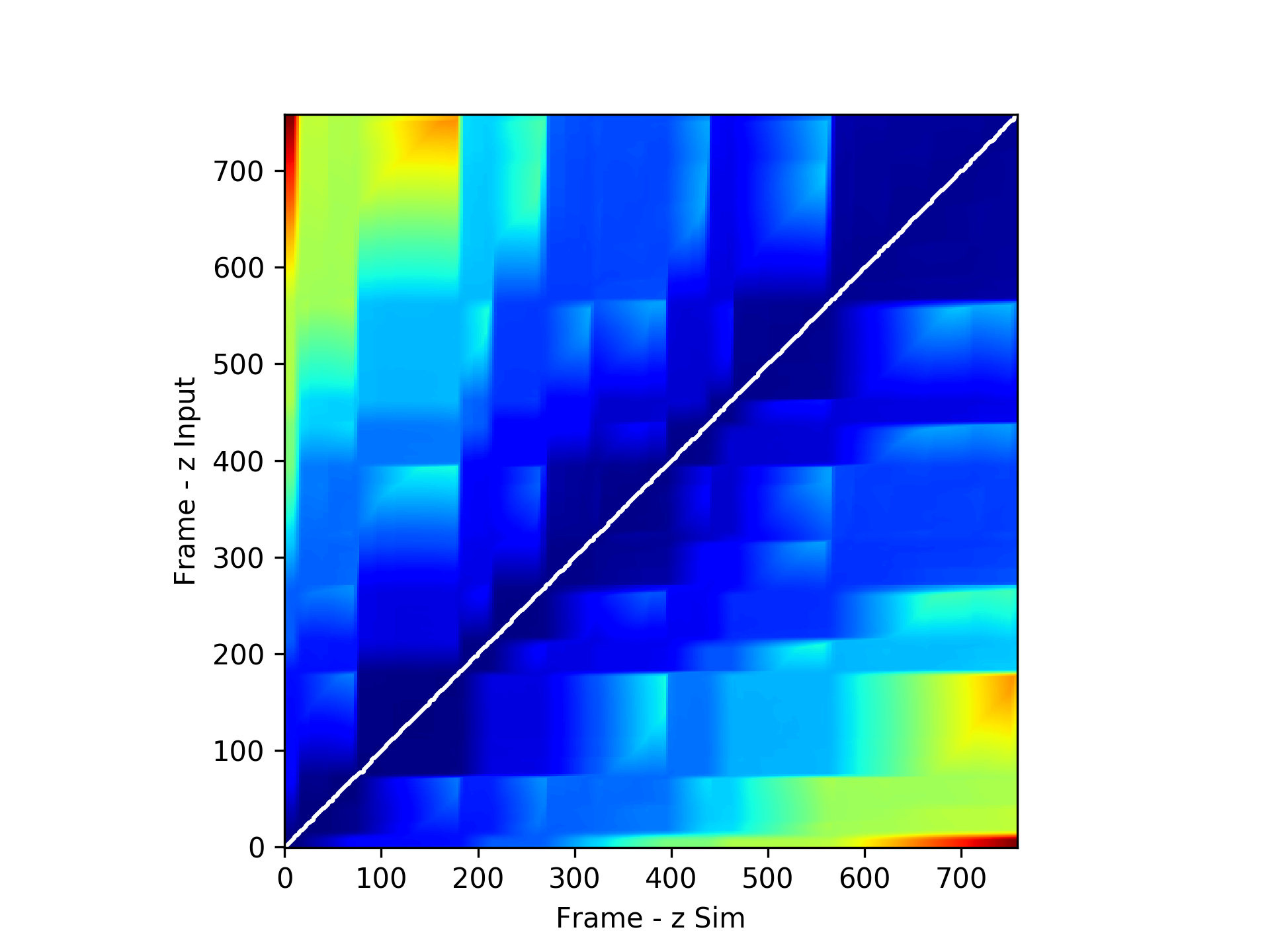}
    \end{subfigure}
    \hspace{-6pt}
    \begin{subfigure}[h]{0.33\columnwidth}
        \centering
        \includegraphics[trim={52pt 0pt 40pt 40pt},clip,width=1.15\columnwidth]{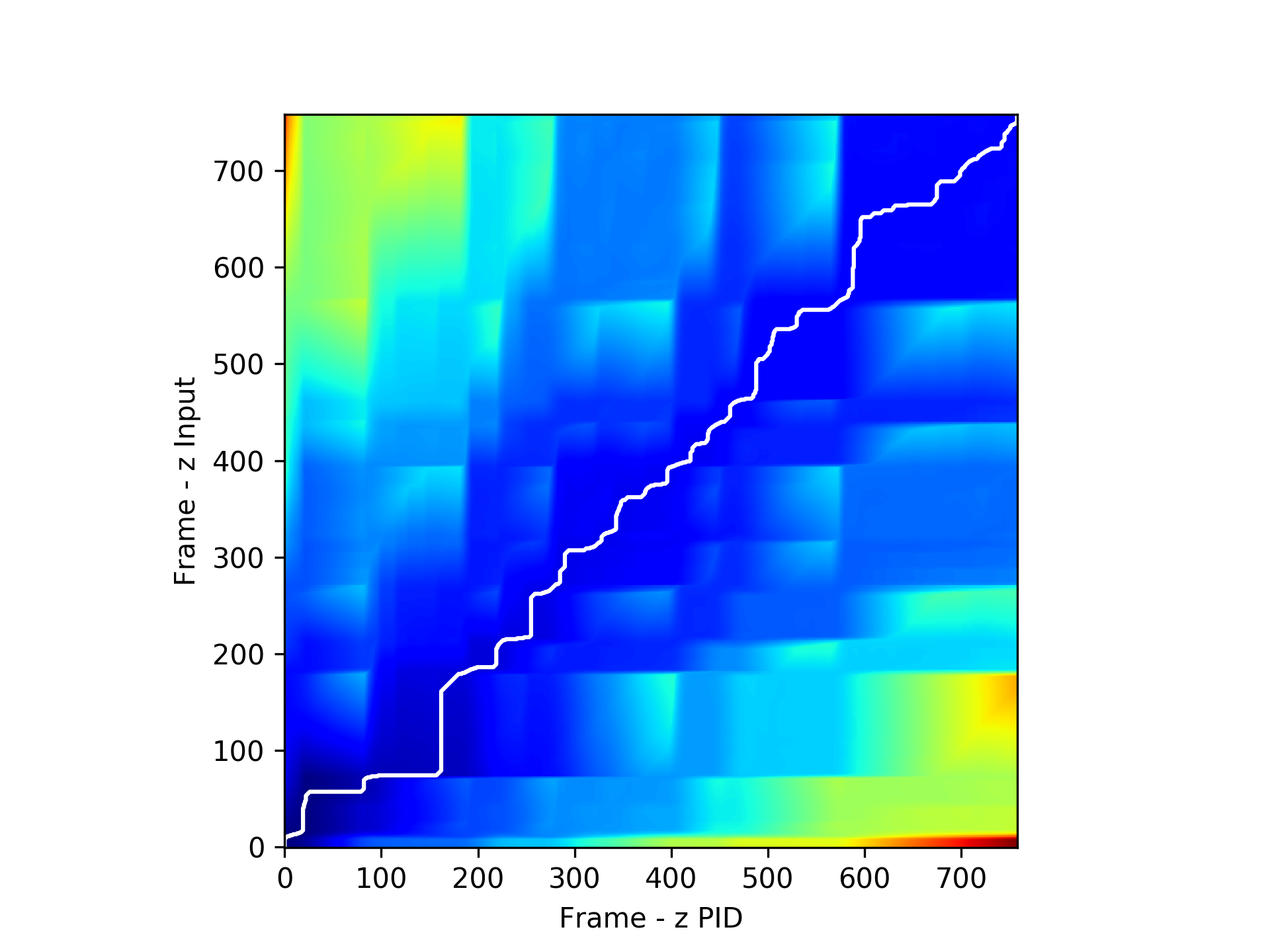}
    \end{subfigure}
    \hspace{-6pt}
    \begin{subfigure}[h]{0.33\columnwidth}
        \centering
        \includegraphics[trim={52pt 0pt 40pt 40pt},clip,width=1.15\columnwidth]{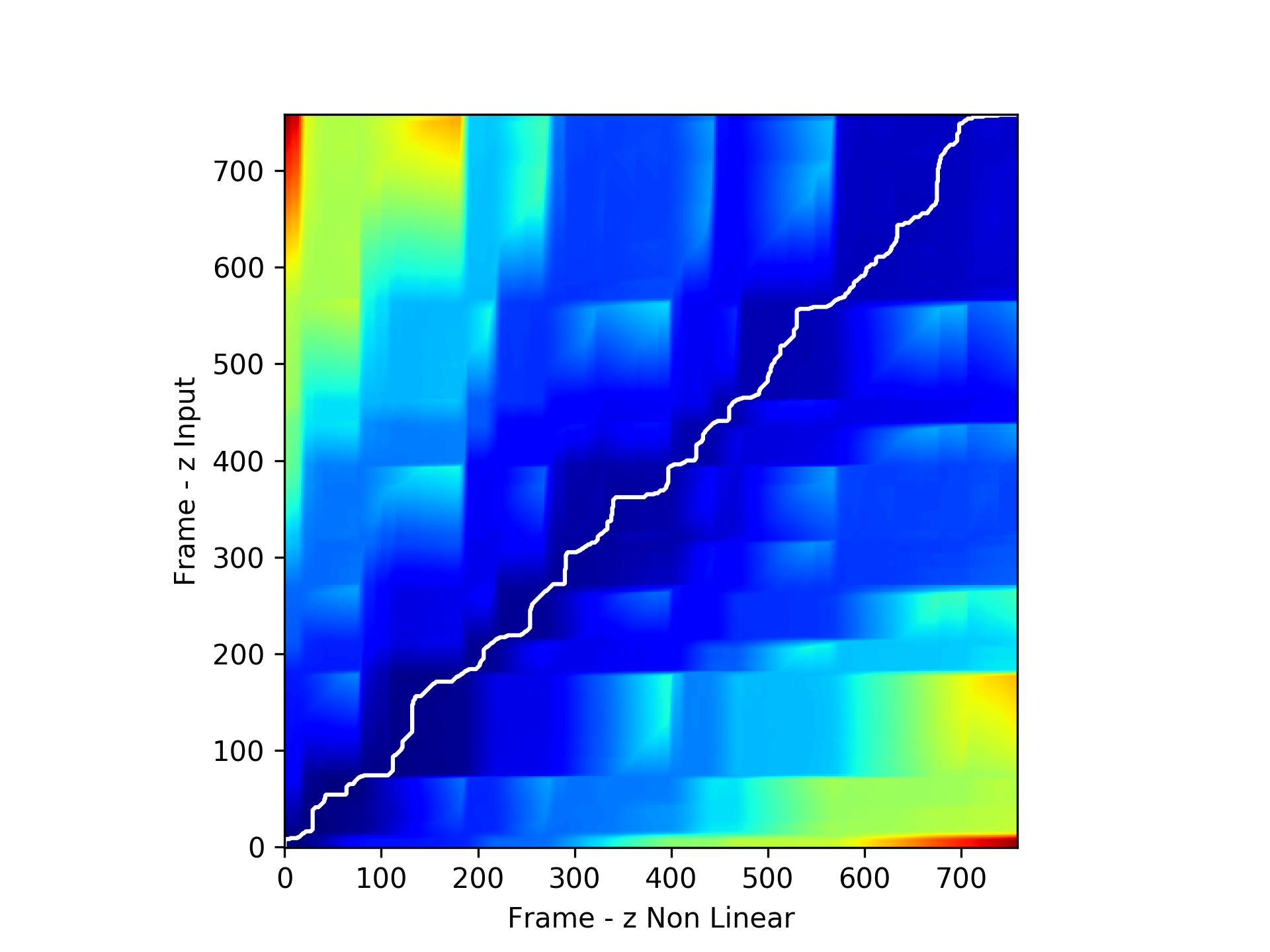}
    \end{subfigure}
 \vspace{-3pt}
\caption{Sample (from S1-T1) Dynamic Time Warping minimum paths in x and z axes for simulation, PID and nonlinear controllers in the physical experiment.}
\label{fig7}
\end{figure}

\begin{figure}[!t]
\vspace{-6pt}
\centering
    \begin{subfigure}[h]{0.3\columnwidth}
        \centering
        \includegraphics[trim={52pt 0pt 40pt 40pt},clip,width=1.15\columnwidth]{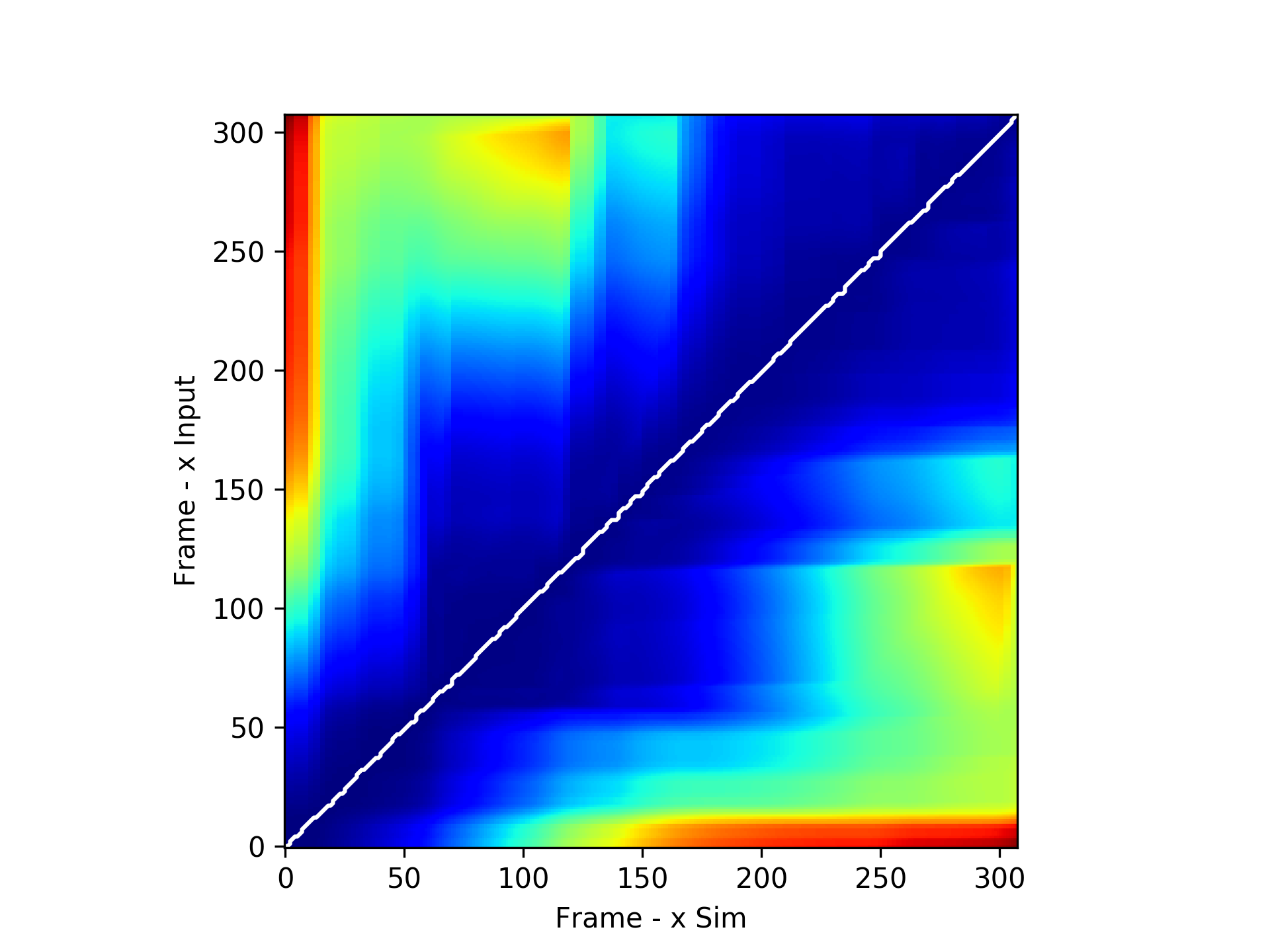}
    \end{subfigure}
    \hspace{-6pt}
    \begin{subfigure}[h]{0.33\columnwidth}
        \centering
        \includegraphics[trim={52pt 0pt 40pt 40pt},clip,width=1.15\columnwidth]{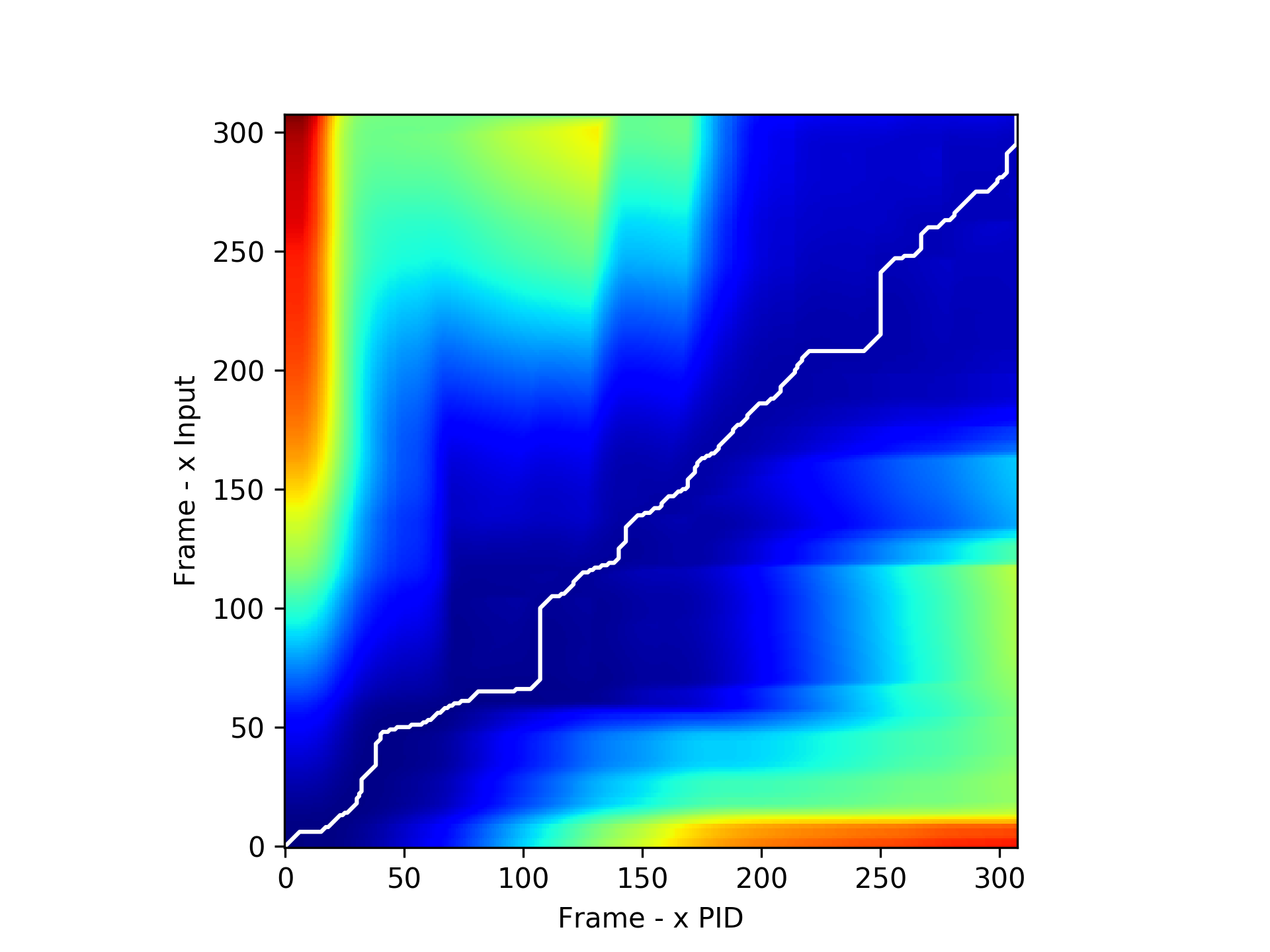}
    \end{subfigure}
    \hspace{-6pt}
    \begin{subfigure}[h]{0.33\columnwidth}
        \centering
        \includegraphics[trim={52pt 0pt 40pt 40pt},clip,width=1.15\columnwidth]{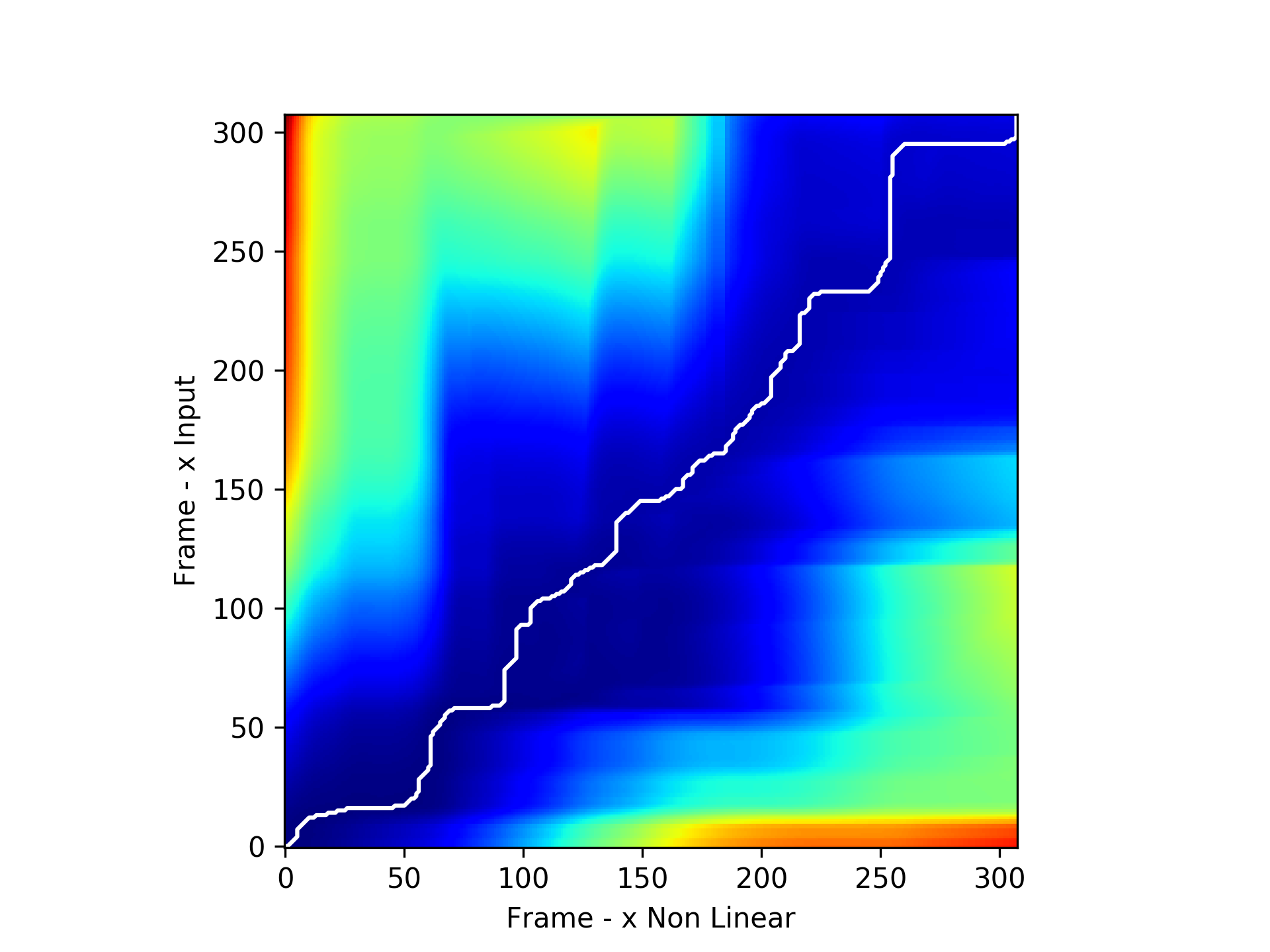}
    \end{subfigure}

    \vspace{3pt}
    \begin{subfigure}[h]{0.33\columnwidth}
        \centering
        \includegraphics[trim={52pt 0pt 40pt 40pt},clip,width=1.15\columnwidth]{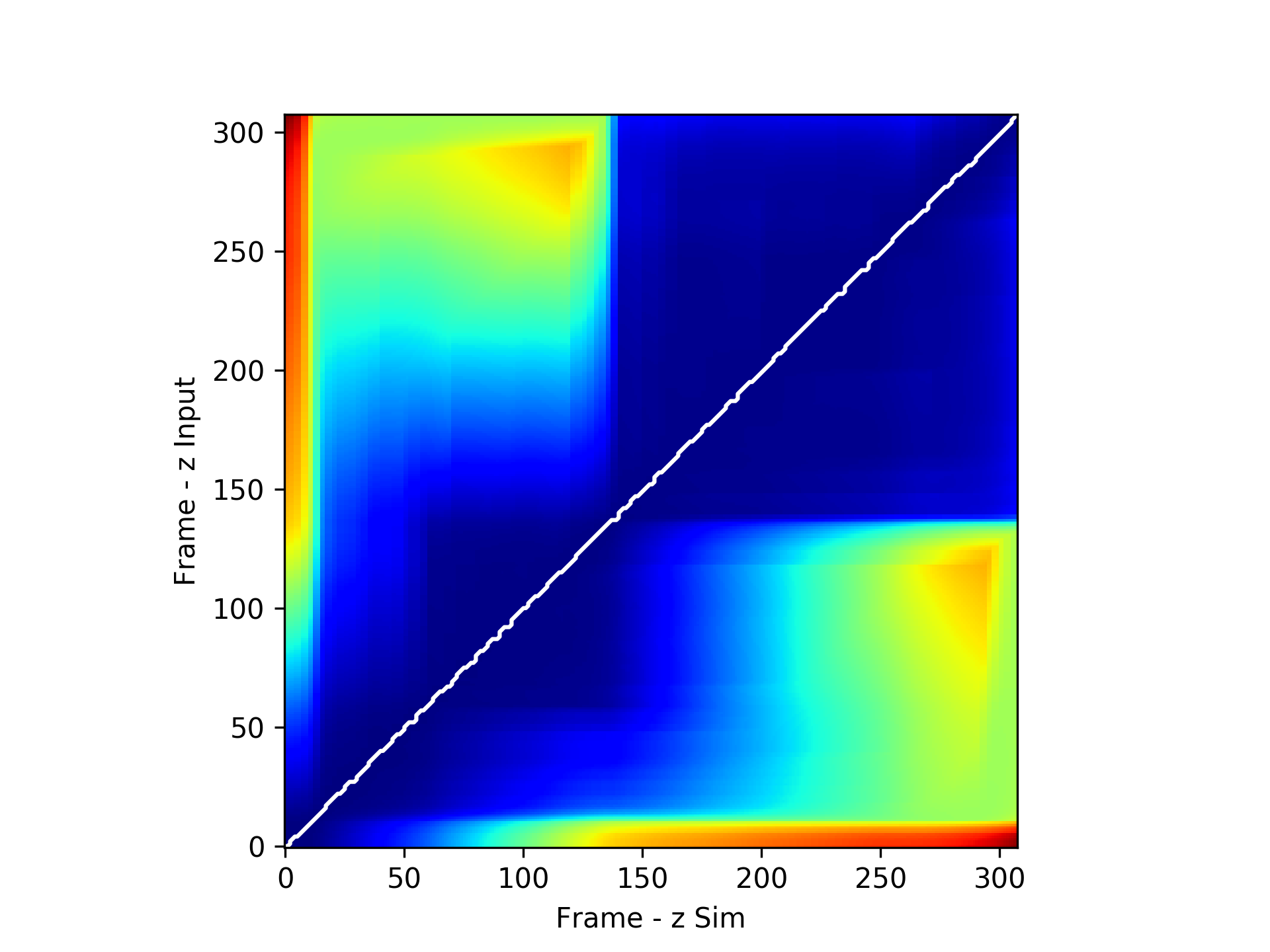}
    \end{subfigure}
    \hspace{-6pt}
    \begin{subfigure}[h]{0.33\columnwidth}
        \centering
        \includegraphics[trim={52pt 0pt 40pt 40pt},clip,width=1.15\columnwidth]{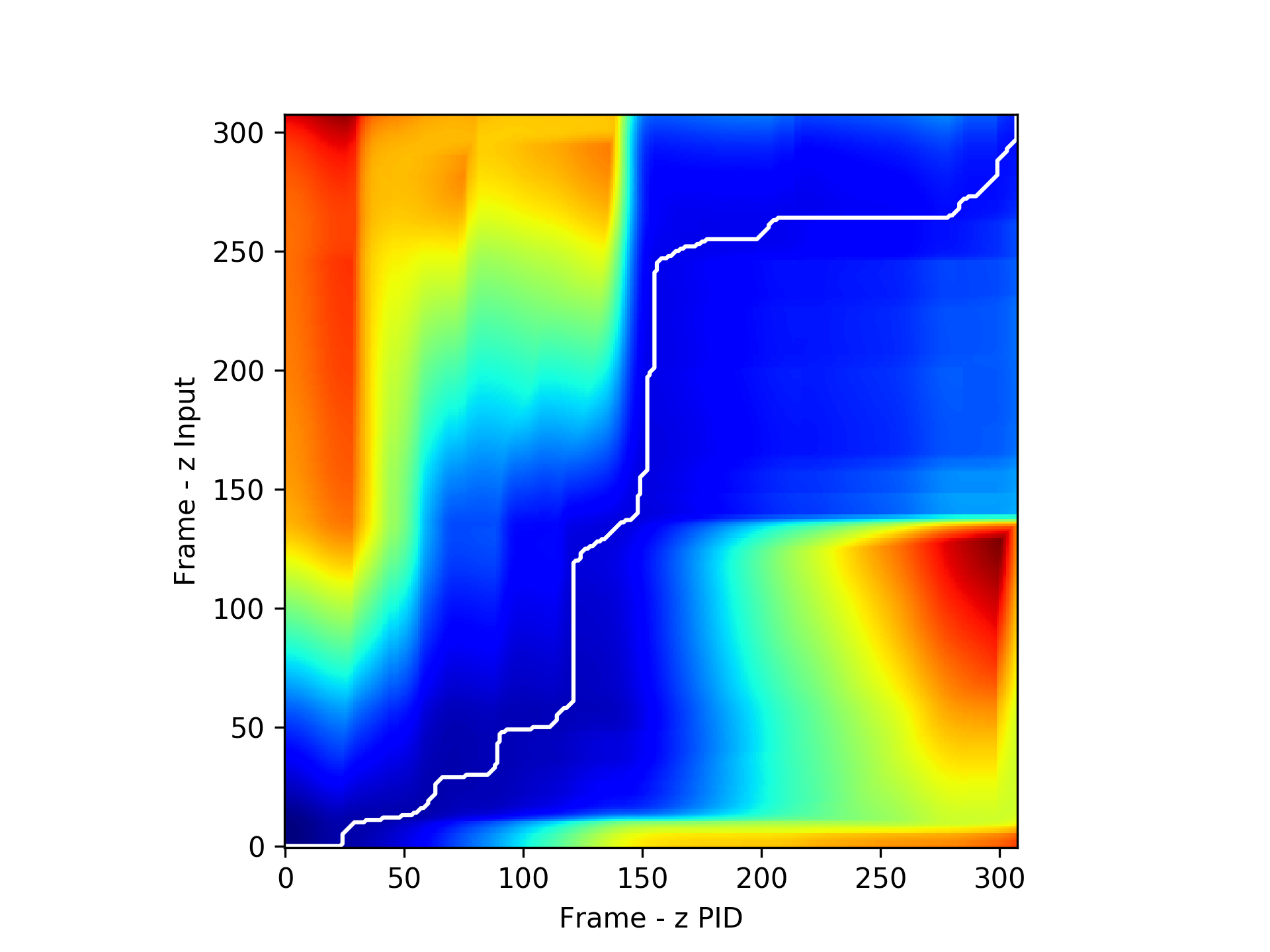}
    \end{subfigure}
    \hspace{-6pt}
    \begin{subfigure}[h]{0.33\columnwidth}
        \centering
        \includegraphics[trim={52pt 0pt 40pt 40pt},clip,width=1.15\columnwidth]{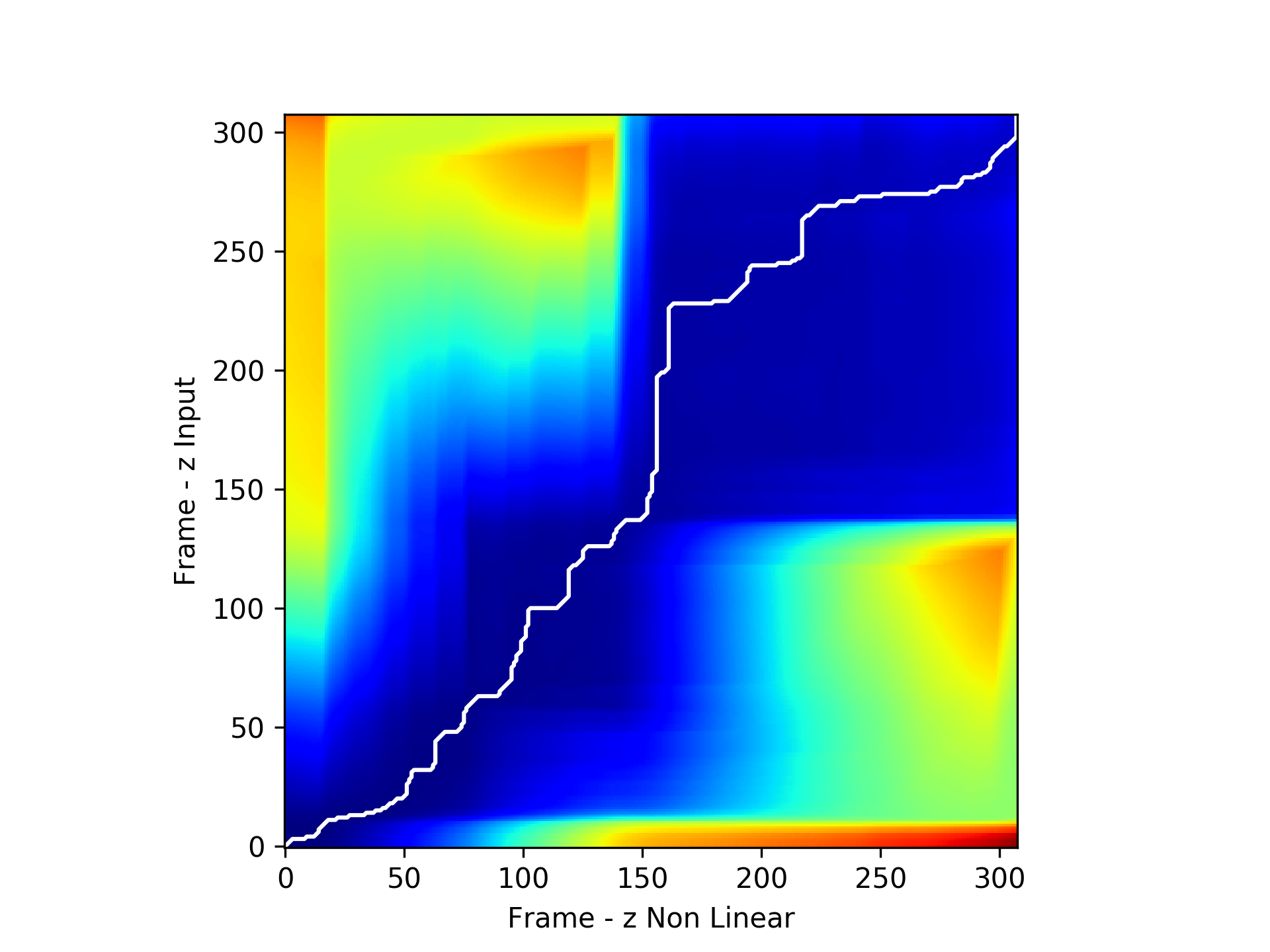}
    \end{subfigure}
 \vspace{-3pt}
\caption{Sample (from S2-T4) Dynamic Time Warping minimum paths in x and z axes for simulation, PID and nonlinear controllers in the physical experiment.}
\label{fig8}
\vspace{-12pt}
\end{figure}

In addition, since the MSE cannot quantify the synchrony between the different time series (which can be readily observed in the signals), we propose to use the Dynamic Time Warping algorithm (DTW) to temporally align the input and output data time series. Each of the line segments encodes an element correspondence (one-to-one or more) between the two discrete signals; Fig.~\ref{fig5} and Fig.~\ref{fig6} demonstrate two examples of the temporal alignment of the input signal and the respective experimental trajectories. 
DTW can provide optimal alignment between time series by computing a minimum distance path between two time series and creating a distance matrix. At each iteration the algorithm computes the minimum distance 
\[d(i, j) = |A_i - B_j| + min(D[i-1, j-1], D[i-1, j], D[i, j-1])\]
where (i, j) are the indices corresponding to elements of each time series,  $A_i$ and $B_i$ are the values at those indices and $D[*,*]$ are elements of the distance matrix. A zero distance value for the DTW minimum path implies absolute matching between the time series~\cite{sakoe1978dynamic}; Fig.~\ref{fig7} and Fig.~\ref{fig8} demonstrate illustrative examples of DTW minimum paths for cases of high and low temporal alignment matching, respectively. A good alignment path does not deviate far from the diagonal. 

With reference to Table~\ref{tab:minPath}, we observe simulation values $d|_x\in [0.23, 4.15]$ in the x axis and $d|_z\in[0.14, 5.73]$ in z axis. Both minimum values (in x and z axis) are observed for S3-T3. In this case the infant retains a leg extension for the majority of the tracked duration and hence the motion is smooth. Both maximum values for the simulation are observed for S5-T2.



Simulation values were significantly lower than those in the physical experiments. 
Regarding the physical experiments, in most of the cases the nonlinear controller minimized the DTW distance in both axes. The minimum values in x axis were found in S2-T4 for the PID controller and S2-T1 for the nonlinear controller. The minimum DTW distance values in z axis were found in S2-T4 for PID and S9-T2 for the nonlinear controller. In the case of S2, the infant appears to retain a knee extension for most of the tracked session, whereas in the case of S9 the infant appears to retain a constant knee flexion throughout the tracked session hence the curve is smooth. The maximum distance values were found in S8-T1, for both axes and both controllers. It is hypothesized that this large value is associated to the larger time duration of the tracked session, which direct effects the calculation of DTW distance values (which is additive). However, graphically, the corresponding minimum path color map the does not deviate far from diagonal.
While the obtained results are overall consistent with the qualitative analysis, additional data would be required to run a proper statistical analysis to further associated observed patters to DTW outputs. 

\section{Conclusion}
In this work we investigated the feasibility of small-scale quadrotor aerial robots to track the evolution of the position of infants' legs motion while kicking and in a supine posture. Infant kicking trajectories were obtained by annotating videos available online. Quadrotor flight was performed in both simulation (to confirm preliminary feasibility testing) as well as via physical experiments. We tested the efficacy of two standard of practice controllers; a linear PID and a nonlinear geometric controller. The ability of the robot to track infant kicking trajectories was evaluated both qualitatively and quantitatively. The MSE was used to assess overall deviation from the input (infant leg trajectory) signals, while the dynamic time warping algorithm was used to quantify the synchrony. Overall we demonstrated that it is possible to track infant kicking trajectories with small-scale quadrotors, and gathered insights regarding areas that require further investigation. Importantly, phase lag between input and output experimental signals was at cases much higher than in simulation, and DTW helps offer a way to indicate when phase lag may increase beyond a desired threshold; this could be used as the means to switch the controller employed by the robot to improve tracking.

In future work we seek to integrate multi-modal sensing (e.g., employ inertial measurement units (IMU) alongside motion capture) to track a larger subset of the robot's state (position, orientation, linear/angular velocities and accelerations). Moreover, we plan to extend the dataset of videos displaying infant kicking motion in supine position and evaluate how principle of machine learning could be used to generate synthetic output trajectories for the aerial robot. 

\bibliographystyle{IEEEtran}
\bibliography{Bibliography/Kouvoutsakis}

\end{document}